\documentclass[preprint,12pt]{elsarticle}

\usepackage{booktabs}
\usepackage{graphicx}
\usepackage{amssymb}
\usepackage{amsmath}
\usepackage{environ}
\usepackage{multirow}
\usepackage{comment}
\usepackage{fixme}
\usepackage[utf8]{inputenc}
\usepackage{algpseudocode}
\usepackage{algorithm}
\usepackage{tikz}
\usepackage{url}
\usepackage{tabularx}
\usepackage{setspace}
\usepackage{algpseudocode}
\usepackage{algorithm}
\usepackage[T1]{fontenc}

\usepackage{tikz}
\usepackage{enumitem}
\usetikzlibrary{arrows.meta}
\usetikzlibrary{tikzmark}
\usetikzlibrary{matrix}
\usepackage[legalpaper, margin=1.2in]{geometry}

\usepackage{tikz}
\usetikzlibrary{positioning}
\definecolor{canaryyellow}{rgb}{1.0, 0.94, 0.0}
\definecolor{brightgreen}{rgb}{0.4, 1.0, 0.0}
\definecolor{jazzberryjam}{rgb}{0.65, 0.04, 0.37}
\usepackage{subcaption}
\usepackage{graphicx}

\fxsetup{
	status=draft,
	author=Marko Djukanović,
	layout=inline,
	theme=color
}
\usepackage{cancel}
\definecolor{fxnote}{rgb}{0.8000,0.0000,0.0000}
\newcommand{\str}[1]{\texttt{#1}}
\usepackage{amssymb}

\journal{Expert Systems with Applications}

\begin{document}

\begin{frontmatter}



\title{A Learning Search Algorithm  for the Restricted Longest Common Subsequence Problem}


\author[label1]{Marko Djukanović\corref{cor1}}
\cortext[cor1]{Corresponding author}
\ead{marko.djukanovic@pmf.unibl.org}
\author[label2]{Jaume Reixach}
\ead{jaume.reixach@iiia.csic.es}
\author[label3]{Ana Nikolikj}
\ead{ana.nikolikj@ijs.si}
\author[label3]{Tome Eftimov}
\ead{time.eftimov@ijs.si}
\author[label4]{Aleksandar Kartelj}
\ead{kartelj@math.rs}
\author[label2]{Christian Blum}
\ead{christian.blum@iiia.csic.es}

\affiliation[label1]{organization={Faculty of Natural Sciences and Mathematics, University of Banja Luka},
            addressline={Mladena Stojanovića 2}, 
            city={Banja Luka},
            postcode={78000}, 
            state={Republic of Srpska},
            country={Bosnia and Herzegovina}}

 \affiliation[label4]{organization={Faculty of Mathematics, University of Belgrade},
	             addressline={Studentski trg 16},
	             city={Belgrade},
	             postcode={104015},
	             country={Serbia}}
	      

 \affiliation[label3]{organization={Jožef Stefan Institute},
	addressline={Jamova cesta 39},
	city={Ljubljana},
	postcode={1000},
	country={Slovenia}}
	

\affiliation[label2]{organization={Artificial Intelligence Research Institute},
	addressline={Campus of the UAB},
	city={Bellaterra},
	postcode={08193},
	country={Spain}}
	            
\begin{abstract}
This paper addresses the Restricted Longest Common Subsequence (RLCS) problem, an extension of the well-known Longest Common Subsequence (LCS) problem. This problem has significant applications in bioinformatics, particularly for identifying similarities and discovering mutual patterns and important motifs among DNA, RNA, and protein sequences. Building on recent advancements in solving this problem through a general search framework, this paper introduces two novel heuristic approaches designed to enhance the search process by steering it towards promising regions in the search space. The first heuristic employs a probabilistic model to evaluate partial solutions during the search process. The second heuristic is based on a neural network model trained offline using a genetic algorithm. A key aspect of this approach is extracting problem-specific features of partial solutions and the complete problem instance. An effective hybrid method, referred to as the learning beam search, is developed by combining the trained neural network model with a beam search framework. An important contribution of this paper is found in the generation of real-world instances where scientific abstracts serve as input strings, and a set of frequently occurring academic words from the literature are used as restricted patterns. Comprehensive experimental evaluations demonstrate the effectiveness of the proposed approaches in solving the RLCS problem. Finally, an empirical explainability analysis is applied to the obtained results. In this way, key feature combinations and their respective contributions to the success or failure of the algorithms across different problem types are identified.

\end{abstract}




\end{frontmatter}


\section{Introduction}\label{sec:introduction}

A string is a finite sequence of characters from an alphabet $\Sigma$. In many programming languages, a string is used as a data structure.  In biology, they represent models for DNA, RNA, and protein sequences. In the field of stringology and bioinformatics, a pivotal task concerns finding meaningful and representative measures of structural similarity between various molecular structures. Among several proposed measures, one that gathered significant attention from a practical and theoretical point of view is finding the  \emph{longest common subsequence} (LCS) for a set of input strings. In this context, a subsequence of a string $s$ is a string obtained by deleting zero or more symbols from $s$, preserving the order of the remaining symbols. Seeking the longest common subsequences has been a subject of intrigue for more than half of a century. The LCS problem is stated as follows. Given a set of input strings $S=\{s_1, \ldots, s_m\}$ over a finite alphabet $\Sigma$, where $m \in \mathbb{N}$ is arbitrary, the aim is to identify a common subsequence of all strings from $S$ with the maximum possible length~\cite{bergroth2000survey}. Apart from bioinformatics, this problem has been important in various fields, such as data compression and text processing~\cite{storer1987data}.  

In the beginning, the focus of scientists has been on developing efficient algorithms for LCS problem cases with fixed $m$, especially for $m=2$. The range of developed algorithms includes dynamic programming (DP), such as the Hirschenberg algorithm, the Hunt-Szymanski algorithm, and the Apostolico-Crochemore algorithm; see~\cite{hirschberg1977algorithms,apostolico1987longest}.  
By the end of the last century, the focus shifted to solving the LCS problem for arbitrary $m>2$. Note that for any fixed $m$, the LCS problem is polynomially solvable by DP. This, however, comes at a considerable price, as the complexity of DP is $O(n^m)$, where $n$ is the length of the longest string in $S$. Thus, DP quickly becomes impractical in general cases.  For arbitrary values of $m$, the problem is known to be $\mathcal{NP}$-hard~\cite{maier1978complexity}. Additionally, it was found that the time complexity of $O(n^{m})$ is likely to be a tight one unless $\mathcal{P}=\mathcal{NP}$. Consequently, the existence of an efficient algorithm for the general LCS problem scenario seems unlikely, yielding the appearance of various heuristic and approximation algorithms in the literature published at the beginning of this century. In particular, approaches based on Beam Search (BS)~\cite{djukanovic2019beam} and hybrid anytime algorithms~\cite{djukanovic2020finding} have been established as the most efficient heuristic and exact approaches, respectively. In parallel with studying the LCS problem, several practically motivated variants of this problem have been introduced and studied. Some of them include the longest arc-preserving common subsequence problem~\cite{lin2002longest,blum2017hybrid}, the constrained LCS problem~\cite{tsai2003constrained,djukanovic2020solving_generalized}, and the shortest common supersequence problem~\cite{mousavi2012enhanced_scs}, among others. \\

This study deals with the \emph{restricted longest common subsequence} (RLCS) problem, initially described by Gotthilf et al.~\cite{gotthilf2010restricted}. In addition to considering an arbitrary set of input strings $S$ over a finite alphabet $\Sigma$, the problem involves an arbitrarily large set of restricted pattern strings $R=\{r_1, \ldots, r_k\}$ over the same alphabet. The objective is to find the longest common subsequence $s$ of the strings in $S$ such that none of the restricted patterns $r_i \in R$ is a subsequence of $s$. In~\cite{gotthilf2010restricted}, the authors show that the RLCS problem is $\mathcal{NP}$-hard even with two input strings and an arbitrary number of restricted patterns. Moreover, they develop a DP approach for general values of $m$ and $k$. In this scenario, they proved that RLCS is fixed-parameter tractable (FTP) when parameterized by the total length of the restricted patterns. In addition, the authors propose two approximation algorithms: one ensures an approximation ratio of $1/|\Sigma|$ 
and the other one guarantees a ratio of 
$(k_{\min}-1) / n_{\min}$, where $k_{\min}$ and $n_{\min}$ represent the lengths of the shortest strings in $R$ and $S$, respectively. \\

Independently of Gotthilf et al.~\cite{gotthilf2010restricted}, Chen and Chao~\cite{chen2011generalized} proposed a DP approach specifically for the RLCS problem with $m=2$ and $k=1$,  which runs in $O(|s_1| \cdot |s_2| \cdot |r_1|)$ time. For this special case of the RLCS problem, Deorowicz and Grabowski~\cite{deorowicz2014subcubic} introduced two asymptotically faster sparse DP algorithms than the conventional dynamic approach. They require subcubic time complexities of $O(|s_1| \cdot |s_2| \cdot |r_1| /\log(|s_1|))$ and $O(|s_1| \cdot |s_2| \cdot |r_1| /\log^{\frac{3}{2}}(|s_1|))$, respectively, by utilizing special internal data structures. Farhana and Rahman~\cite{farhana2015constrained} proposed a finite automata-based approach for the general RLCS problem that runs in  $O(|\Sigma| (\mathcal{R} + m) + nm + |\Sigma| \mathcal{R} n^k)$ time, where $\mathcal{R}=O(n^m)$ denotes the size of the resulting automaton. Recently, Djukanovic et al.~\cite{djukanovic_rlcs_search_2024} proposed a general search framework to solve the RLCS problem. In particular, an exact A$^*$ search and a heuristic BS approach are derived from this framework. These algorithms are currently the state of the art in terms of exact and heuristic solvers. However, the current literature has an obvious limitation: all algorithms were tested on the instances with limited sizes, that is, with at most five input strings. The same holds for the set of pattern strings.  Therefore, the effectiveness and limitations of the existing approaches for solving the tackled instances remained weakly explored, despite their superiority in contrast to the remaining competitor approaches from the literature.  \\

This paper is a substantial extension of our earlier conference paper published in~\cite{djukanovic_rlcs_search_2024}. The main contributions are as follows:
\begin{itemize}
	\item A novel heuristic guiding function is proposed based on an enhanced probability model for the LCS problem, incorporating an additional tie-breaking mechanism.
	
	\item  A new learning-based BS approach is introduced. This technique makes use of a neural network that is trained offline using a carefully selected set of both instance-specific and global features of RLCS instances. These features are fed into the input layer of the network, which then guides the BS as a heuristic.
	
	\item A new dataset consisting of 820 larger random instances is generated to evaluate the scalability of the proposed search algorithms. Additionally, a new set of real-world instances is created, where input strings are scientific paper abstracts, and restricted strings are drawn from the 60 most frequent academic terms used in scientific writing.
	
	\item A comprehensive explainability analysis is conducted to assess the performance of all approaches, focusing on various low-level problem features.
	
	\item Extensive experimental results show that the new learning-based BS approach achieves state-of-the-art performance on randomly generated instances and delivers the best overall results on the real-world benchmark set. The statistical findings are further enhanced with an empirical explainability analysis, which highlights the critical feature combinations responsible for the overall performance of each approach.
\end{itemize}

\subsection{Preliminaries}
Before we dive into more complex issues, let us prepare the ground with a few definitions and notation. The length of a string $s$ is denoted by $|s|$. By $s[i], 1 \leq i \leq |s|$, the $i$-th character of string $s$ is referred to. Note that the position of the leading character is indexed with 1. For two integers $i, j \leq |s|$, $s[i, j]$ is the contiguous part of the string $s$ that begins with the character at position $i$ and ends with the character at position $j$; if $i=j$, the single-character string $s[i] = s[i, i]$ is given; finally, if $i > j$, $s[i,j]$ refers to the empty string $\varepsilon$. For a left position vector $p^{L}=(p^L_1, \ldots , p^L_m), \ 1 \leq p^L_i \leq |s_i|, \ i=1, \ldots , m$, we denote by $S[p^{L}]$ the set of suffix input strings associated with the respective coordinates of this vector, i.e., $S[p^{L}]:= \{ s_i[p^L_i, |s_i|] \mid i=1, \ldots, m\}$.  Given string $s$ and letter $a$, by  $|s|_a$, we denote the number of times letter $a$ appears in string $s$. By $s^{rev}$, we denote the reversed string of string $s$. Finally,  by $Succ(x)_{i,a}$, we denote the smallest index ($y$) greater or equal to $x$ so that $s_i[y]=a$, $i=1,\ldots, m$; if no such position exists, $Succ(x)_{i,a}:=|s_i|+1$. 

A complete RLCS problem instance is denoted as $(S, R, \Sigma)$, where $S$ contains the input strings, $R$ the restricted pattern strings, and $\Sigma$ is the finite alphabet. For two integer vectors $\textbf{p} \in \mathbb{N}^m$ and $\textbf{q} \in \mathbb{N}^k$, a sub-problem (sub-instance) of the initial problem instance concerning these two (left) positional vectors is denoted by $(S[\textbf{p}], R[\textbf{q}])$. From now onwards, by $n$ we denote the length of the longest input string from $S$.  
\\
The remaining sections of the work are organized as follows. Section~\ref{sec:search_algorithms} details the general search framework for solving the RLCS problem. In particular, an A$^*$ search and a BS algorithm are derived and explained in sufficient detail. Additionally, two classical heuristic functions are explained for the tackled problem. Section~\ref{sec:learning-bs} introduces a learning BS approach. The exhaustive experimental evaluation is provided in Section~\ref{sec:experimental_evaluation}. Section~\ref{sec:ela} is devoted to providing a deeper understanding of the algorithms' performances by performing an explainability analysis. The paper finally concludes with Section~\ref{sec:future_work} along with directions to future research.

\section{Search Approaches for the RLCS Problem} \label{sec:search_algorithms}

The general search framework for the RLCS problem was proposed in~\cite{djukanovic_rlcs_search_2024}. To ensure completeness of the present paper, this search framework is explained in the following section, mainly following the notation introduced in the aforementioned paper.  Afterward, we show two derivatives of this framework: an A$^*$ search approach and BS. 

\subsection{The General Search Scheme}

The state graph represents the environment of our proposed algorithms. Its inner nodes represent partial solutions, while its leaf nodes represent complete solutions. Moreover, edges between nodes represent extensions of partial solutions. The state graph $G=(V,E)$ of an RLCS problem instance $(S,R)$ is defined as follows. 

We say that a partial solution---that is, a common subsequence $s^v$ of the strings in $S$ that does not contain any string from $R$ as subsequence---\textit{induces} a state graph node $v=(p^{L,v}, l^v, u^v) \in V$ if:
\begin{itemize}
	\item $|s^v|=u^v$
	\item $s^v$ is a subsequence of all $s_i[1, p^{L,v}_i-1], i \in \{1, \ldots, m\}$ and $p^{L,v}_i - 1$ is the smallest index that satisfies this property.
	\item $s^v$ contains none of the prefix strings $r_j[1, l^v_j ] $ as its subsequence whereas $r_j[1, l^v_j-1], j \in \{1, \ldots, k\}$ are all included.
\end{itemize}

There is an edge (transition) between nodes $v_1= (p^{L,v_1}, l^{v_1}, u^{v_1})$ and $v_2=(p^{L,{v_2}}, l^{v_2}, u^{v_2})$ labelled with a letter $a \in \Sigma$, denoted by $t(v_1, v_2)=a$, if:
\begin{itemize}
	\item $u^{v_1}+1 = u^{v_2}$
	\item The partial solution inducing node $v_2$ is obtained by appending the letter $a$ to the partial solution inducing node $v_1$.
\end{itemize}

Each edge of the state graph of an RLCS problem instance has weight one and (as mentioned above) a label denoting the letter used to extend the respective partial solution. \\

\textit{Node extension}. The process of determining the successor (child) nodes of a node $v$ is called \textit{extending} $v$. For doing so, we identify those letters that can feasibly extend the partial solution $s^v$ represented by $v$. This procedure consists of three steps. First, all letters that occur in each string from $S[p^{L,v}]$ are identified. Second, a letter that causes a violation of the restrictions is removed. This happens if one of the restricted patterns $r_i\in R$ becomes a subsequence of the partial solution generated by extending $s^v$ with that letter. Third, dominated letters are omitted from consideration. Letter $a$ is said to \textit{dominate} letter $b$ (i.e., $b$ is dominated by $a$) if $Succ[p^{L,v}i]_{i, a} \leq Succ[p^{L,v}i]_{i, b}$ for every $i \in \{ 1, \ldots, m\}$ and $r_j[ l^v_j] \notin \{a, b\}$ for all $j \in \{1, \ldots ,k\}$. The set of non-dominated feasible letters to extend the partial solution of a node $v$ is denoted by $\Sigma^{nd}_v$. Now, for a letter $a \in \Sigma^{nd}_v$, the corresponding successor node $w=(p^{L,{w}}, l^{w}, u^{w})$ of $v$ is constructed in the following way.
\begin{itemize}
	\item $ u^{w}= u^{v}+1$, because the partial solution of node $v$ derives the partial solution of node $w$ by appending the letter $a$ to it: $s^{w}=s^{v}\cdot a$.
	
	\item $l^{w}_j=l^{v}_j + 1$ if $r_j[ l^{v}] = a $ or $l^{w}_j=l^{v}_j $ otherwise.
	
	\item For the (left) position vectors we have $p^{L,{w}}_i = Succ[ p^{L,v}_i ]_{i, a} + 1$ where $Succ(x)_{i,a}$ represents the smallest index greater or equal to $x$ so that $s_i[y]=a$.
\end{itemize}

Note that the data structure \emph{Succ} is preprocessed before the construction of the RLCS state graph has started. This ensures finding suitable position vectors of a child node in $O(m)$ time. \\

The \textit{root} (initial) node is given by $r=((1, \ldots, 1), (1, \ldots, 1), 0)$ and it corresponds to the empty solution $s^r=\varepsilon$ that is trivially feasible and induces the complete problem instance $(S,R)$. A node $v$ is \emph{complete} if $\Sigma^{nd}_v = \emptyset$, that is, if it does not have any child nodes (successors). Note that (partial) solutions that induce complete nodes are candidates for optimal solutions. In this context, optimal solutions are endpoints of the longest paths from the root node $r$ to complete nodes.  
Since the RLCS problem is $\mathcal{NP}$-hard, the entire state graph is generally infeasible to create as its size grows exponentially with the instance size. Consequently, the algorithms proposed generate and visit nodes on the fly, employing a set of intelligent decisions towards prioritizing the exploration of more promising nodes first. How these decisions are taken is explained in the following sections. \\

\begin{figure}[!t]
	\centering
	\scalebox{0.60}{
		\begin{tikzpicture}[>=latex,xscale=1.60, yscale=1.15, every node/.style={transform shape}]
			\begin{scope}[every node/.style={fill=blue!20,rounded corners,thick,draw}]
				\node (A) at (4, 4.5){$\textbf{\small{((1,1),(1,1),0)}}$};
				\node (B) at (2, 3)  {$\textbf{\small{((2,3),(1,2),1)}}$};
				\node (B1) at (0, 3)  {$\textbf{\small{((4,6),(1,1),1)}}$};
				\node (C1) at (-0.0, 1)  {$\textbf{\small{((6,7),(2,1),2)}}$};
				\node (C) at (6, 3)  {$\textbf{\small{((3,2), (2,1), 1)}}$};
				\node (D2) at (9, 3)  {$ \textbf{\small{\xcancel{((8,8), (1,1), 1)}}}$};
				\node[fill=white] (D3) at (9, 4)  {$\str{G}$ dominated by $\str{A}$};
				\node (D) at (1.95, 1.0){$\textbf{\small{((3,4),(2,2), 2)}}$};
				\node[fill=lightgray] (D1) at (-1.9, 1)  {$\textbf{\small{((7,9),(1,2),2)}}$};
				\node[fill=lightgray] (DN) at (-4.0, 1)  {$\textbf{\small{((8,8),(1,1),2)}}$};
				\node[fill=lightgray] (DN2) at (-4.0, -1)  {$\textbf{\small{((9,9),(1, 2),3)}}$};
				
				
				\node (DNN) at (4.0, 1.0){$\textbf{\small{((8, 8),(1,2), 2)}}$};

				\node (E) at (3, -1) {$\textbf{\small{((6,5), (2, 2),3)}}$};
				\node[fill=lightgray] (E1) at (0, -1)  {$\textbf{\small{((7,9),(3,2),3)}}$};
				\node[fill=lightgray] (G) at (3, -3) {$\textbf{\small{((7,9),(3,2),4)}}$};
				\node[] (GN) at (1, -3) {$\textbf{\small{((8,8),(2,2),4)}}$};
				
				
				\node (L) at (6.0, 1) {$\textbf{\small{((4,6),(2,1),2)}}$};
				\node (M) at (10.0, 1) {$\textbf{\small{((6,4),(2,1),2)}}$};
				\node[] (N) at (8.0, 1) {$\textbf{\small{((7,3),(3,2),2)}}$};
				
				\node (R) at (5.1, -1) {$\textbf{\small{((6,7),(2,1),3)}}$};
				\node[fill=lightgray] (RN) at (7.7, -1) {$\textbf{\small{((8, 8),(3,2), 3)}}$};
				
				\node[] (RNN) at (10.0, -1) {$\textbf{\small{((8, 8),(2,1), 3)}}$};
				
				\node[fill=lightgray] (RNNN) at (10.0, -3) {$\textbf{\small{((9, 9),(3,2), 4)}}$};
				
				\node[] (GNNNNNN) at (5.1, -3) {$\textbf{\small{((8,8),(2,1),4)}}$};

				\node[fill=lightgray] (GNNNN) at (3, -3) {$\textbf{\small{((9,9),(3,2),4)}}$};
				
				\node[fill=lightgray] (GNNNNN) at (5.1, -5) {$\textbf{\small{((9,9),(3,2),5)}}$};

			\end{scope}
			\begin{scope}[>={Stealth[red,scale=0.8]},
				every node/.style={fill=white,circle},
				every edge/.style={draw=red,thick,line width=0.8pt}]
				\path [->] (A) edge[color=blue, >=stealth=blue, line width=2.5pt] node {$\str{T}$} (B);
				\path [->] (A) edge[ ] node {$\str{A}$} (B1);
				
				\path [->] (A) edge[ ] node {$\str{G}$} (D2);
				
				\path [->] (B1) edge[ ] node {$\str{C}$} (C1);
				\path [->] (B1) edge[ ] node {$\str{T}$} (D1);
				\path [->] (B) edge[bend right=-12 ] node {$\str{T}$} (D1);
				\path [->] (D) edge[bend right=-12 ] node {$\str{T}$} (E1);
				\path [->] (M) edge[bend right=4] node {$\str{T}$} (E1);

				\path [->] (C1) edge[ bend right=-10 ] node {$\str{T}$} (E1);  
				\path [->] (A) edge[color=blue, >=stealth=blue, line width=2.5pt] node {$\str{C}$} (C);
				\path [->] (B) edge[color=blue, >=stealth=blue, line width=2.5pt] node {$\str{C}$} (D);
				\path [->] (D) edge[color=blue, >=stealth=blue, line width=2.5pt] node {$\str{C}$} (E);
				\path [->] (E) edge[] node {$\str{T}$} (G);
				\path [->] (C) edge[color=blue, >=stealth=blue, line width=2.5pt] node {$\str{A}$} (L);
				\path [->] (C) edge node {$\str{C}$} (M);
				\path [->] (C) edge node {$\str{T}$} (N);
				
				\path [->] (B1) edge node {$\str{G}$} (DN);
				\path [->] (DN) edge node {$\str{T}$} (DN2);
				\path [->] (B) edge node {$\str{G}$} (DNN);
				
				\path [->] (DNN) edge[bend right=2 ] node {$\str{T}$} (DN2);
				
				\path [->] (C1) edge[bend right=3 ] node {$\str{G}$} (RNN);
				\path [->] (N) edge[bend right=0 ] node {$\str{G}$} (RN);
				\path [->] (M) edge node {$\str{G}$} (RNN);
				\path [->] (RNN)  edge[] node {$\str{T}$} (RNNN);
				\path [->] (E)  edge[color=blue, line width=2.5pt, >=stealth=blue] node {$\str{G}$} (GN);
				\path [->] (GN)  edge[color=blue, line width=2.5pt, >=stealth=blue] node {$\str{T}$} (GNNNNN);
				\path [->] (R) edge[color=blue, line width=2.5pt, >=stealth=blue] node {$\str{G}$} (GNNNNNN);       
				\path [->] (GNNNNNN) edge[color=blue, line width=2.5pt, >=stealth=blue] node {$\str{T}$} (GNNNNN);  
				\path [->] (L) edge[color=blue, >=stealth=blue, line width=2.5pt] node {$\str{C}$} (R);
			\end{scope}
	\end{tikzpicture}}
	\caption[ Example showing the full state graph for a problem instance.]{Example of the full state graph in the form of a directed acyclic graph for the problem instance $(S=\{s_1=\str{TCAACTGT}, s_2=\str{CTCCACGT}\},\ R=\{r_1=\str{CTT}, r_2=\str{TA}\}$). 
		It contains eight complete nodes (light grey background). The two paths from  $((9, 9), (3, 2), 5)$ to the root node (in blue) are the longest paths in the graph. Hence, they represent two optimal solutions for this problem instance, i.e., $\str{TCCGT}$ and $\str{CACGT}$, respectively.}\label{fig:example_state_graph}    
\end{figure}
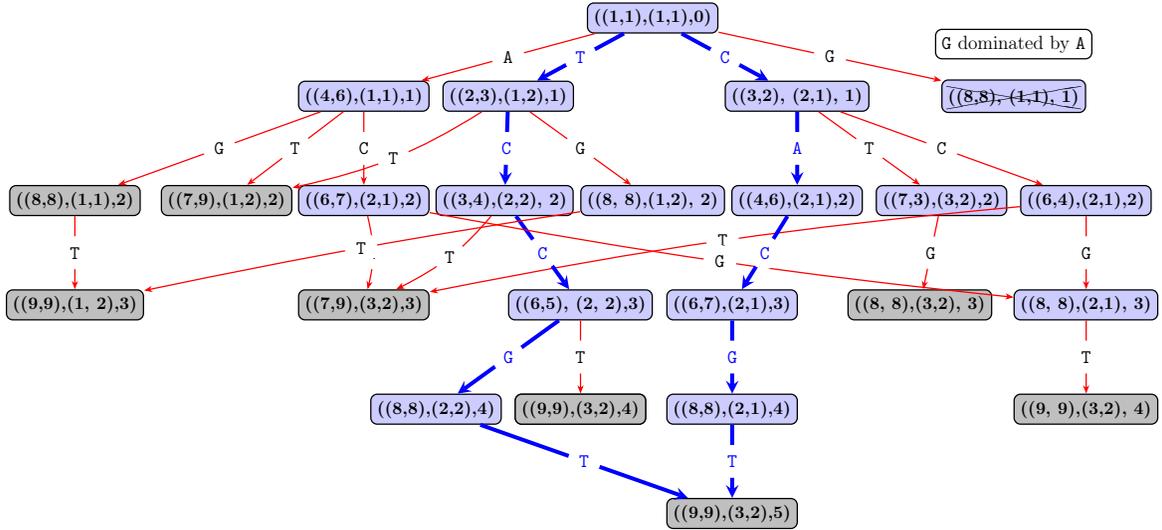

This section concludes by showing the complete state graph of an example problem instance in Figure~\ref{fig:example_state_graph}. Note that the root node $r$ has three child nodes. This is because the (possible) partial solution \str{G}---corresponding to node $((8, 8),(1,1),1)$, which is a feasible extension of the root node---is dominated by \str{A} and therefore would generate a provenly suboptimal path. Hence, it has been omitted from the search. Note the same is encountered with node $((3, 4),(2, 2), 2)$ where \str{G} is dominated by \str{C}.

 \subsection{A$^*$ Search Algorithm}\label{sec:a_star} 

A$^*$ search~\cite{hart1968formal} is an exact, informed search algorithm that follows the best-first search strategy of node processing. It is the most widely used path-finding algorithm; it has been applied to solve many problems, including video games, string matching and parsing, knapsack problems, and others~\cite{kapi2020review,sbihi2007best}. Its core principle is to expand the most promising node at each iteration. To evaluate a node $v$, a scoring function $f(v) = g(v) + h(v)$ is utilized. By assuming the goal is to find the longest path, as in the case of the RLCS problem, functions $g()$ and $h()$ are defined as follows:
\begin{itemize}
	\item $g(v)$ is the current longest path from the root $r$ to $v$.
	\item $h(v)$ is a heuristic estimation of the length of the longest path from $v$ to a complete (goal) node.
\end{itemize}

Note that A$^*$ search works on a dynamically created directed acyclic graph and in practice rarely examines all nodes. It has the advantageous ability to merge multiple nodes into one, which, as explained below, leads to considerable memory savings. 

To set up an efficient A$^*$ search for the RLCS problem, two crucial data structures are leveraged:

\begin{itemize}
	\item A hash map $N$ with keys represented by pairs $(p^{L, v}, l^v)$, with the corresponding value as the longest partial solution that induces a node with these vectors, referring the sub-instance $(S[p^{L, v}], R[l^v])$ to be solved. This data structure efficiently checks whether or not a node with the same key values has already been visited during the search.
	\item A priority queue $Q$ that contains open (not yet expanded) nodes, prioritized on their $f$-values. This structure facilitates the efficient retrieval of the most promising node.
\end{itemize}

The efficacy of A$^*$ search is usually related to the tightness of the heuristic $h()$ function. Therefore, in our application to the RLCS problem, we use the tightest known upper bound (UB) for the LCS problem, adapted to the RLCS problem; the details are explained in Section~\ref{sec:upper_bounds}. When multiple nodes with the same $f$-value exist, ties are resolved by favoring those with a higher $u^v$ value. \\

The pseudocode of our A$^*$ approach is presented in Algorithm~\ref{alg:a_star_rlcs}. The algorithm begins by initializing the root node $r$, which is subsequently added to both the explored set $N$ and the priority queue $Q$. At each iteration, the most promising node $v$ is selected from the top of the priority queue $Q$. If node $v$ represents a complete solution, the corresponding solution is reconstructed by tracing the path from $v$ back to the root node $r$, reading the letters along the transitions, and returning the provenly optimal solution. If $v$ is not a complete node, the algorithm expands it by generating its child nodes. For each child node $v'$, the tuple $(p^{L,v'}, l^{v'})$ is checked to see if it already exists in $N$. If not, $v'$ is added to both $N$ and $Q$. If it does exist, the algorithm checks whether a better path from the root node $r$ to the node associated with $(p^{L,v'}, l^{v'})$ has been discovered. If so, the corresponding information in $N$ is updated, and the priority of $v'$ in $Q$ is adjusted. If not, $v'$ is considered irrelevant and skipped from further consideration. The algorithm continues iterating until either time or memory resources are exhausted. A mechanism is employed throughout the process to track the best complete solution found so far, denoted as $s_{best}$, at each iteration.

\begin{algorithm}[!t]
	\caption{A$^*$ search for the RLCS problem.}\label{alg:a_star_rlcs}
	\begin{algorithmic}[1]
		\Statex \textbf{Data structures:} $N$: the hash map containing generated combinations of position vectors $(p^{L, v}, l^v)$ where the value $N[ (p^{L, v}, l^v) ]$  represents the length of the currently longest path for this combination of vectors; $Q$: priority queue with all not yet processed nodes;
		\Statex \textbf{Input}: problem instance $(S, R)$ of dimension $m \times k$; UB: admissible upper bound function
		\State $ p^{L,v} \gets (1,\ldots,1)$ (length $m$), $l^v \gets (1,\ldots,1)$ (length $k$)
		\State $r \gets ( p^{L,v}, l^v, 0)$
		\State $N[ (p^{L, v}, l^v) ] \gets 0$
		\State $Q \gets \{r\}$
		\State $s_{best} \gets \varepsilon$
		
		\While{\emph{time $\wedge$ memory limit} are not exceeded $\wedge$ $Q\neq \emptyset$}
		\State $v \gets Q.pop()$
		\If{$v$ is complete}
		\State {\bf return} proven optimal solution $s_{best}$  
		\Else
		\State $\Sigma_v^{\mathrm{nd}}\gets$  Determine the feasible non-dominated extensions of node $v$
		\For{$c \in \Sigma_v^{\mathrm{nd}}$}
		\State Generate node $v'$ from node $v$ via extension by letter $c$
		\If{$(p^{L, v'}, l^{v'}) \in N$}
		\If{$N[ (p^{L, v'}, l^{v'}) ] > u^{v}$} \hspace{0.5cm}// a better path found
		\State $N[ (p^{L, v'}, l^{v'}) ] \gets u^{v'}= u^{v}+1$
		\State Update priority value of node $v$ in $Q$
		\EndIf
		\Else \hspace{0.5cm} // a new node examined
		\State $f_{v'} \gets u^{v'} + \text{UB}(v')$ \#UB provided by Eq. (3)
		\State Insert $v'$ into $Q$ with priority value $f_{v'}$
		\State Insert $v'$ into $N$
		\EndIf
		\If{$u^{v'} > |s_{best}|$} \hspace{0.3cm}\# keep track of the best-found solution
		\State $s_{best} \gets$ Derive the solution represented by $v'$
		\EndIf
		\EndFor
		\EndIf
		\EndWhile
		\State {\bf return} $ s_{best}$
	\end{algorithmic}
\end{algorithm}

\subsection{Beam Search Algorithm} \label{sec:bs}

Beam Search (BS)~\cite{lowerre1976harpy} is a heuristic tree-search algorithm that works in a ``breadth-first-search'' (BFS) manner, expanding up to $\beta>0$ best nodes at each tree level. Parameter $\beta$ ensures the size of the BS tree remains polynomial with the instance size, which makes this method robust in providing high-quality solutions to complex problems. The value of $\beta$ controls the trade-off between greediness and completeness. BS is widely used in many fields such as packing~\cite{akeb2009beam}, scheduling~\cite{sabuncuoglu1999job}, and bioinformatics~\cite{carlson2006beam}, among others.

The effectiveness of BS is not only governed by the beam width $\beta$, but also heavily relies on the heuristic function $h()$, which evaluates the quality of each node during the search. The choice of $h()$ is typically problem-specific and sensitive to the characteristics of the problem instances. In the context of our application, we utilize the following heuristic functions: ($i$) the same UB as used in the A$^*$ search outlined above; ($ii$) the probability-based guidance introduced in Section~\ref{sec:prob-based-guidance}; and ($iii$) information derived from a trained neural network as detailed in Section~\ref{sec:learning-bs}.

The BS approach for the RLCS problem is outlined in Algorithm~\ref{alg:bs_rlcs}. The algorithm begins by generating the root node $r$, which is then added to the initial beam $B$ (i.e., $B = {r}$), and initializing the best solution $s_{best}$ to an empty string $\varepsilon$. In the main loop of BS all nodes from the current beam $B$ are expanded in all possible ways, producing a set of child nodes stored in set $V_{ext}$. These child nodes are then sorted in descending order based on their $h()$-values. The top $\beta$ nodes from $V_{ext}$ are selected to form the beam $B$ for the next entry into the algorithms' main loop. This process is repeated until the beam $B$ becomes empty, at which point the algorithm terminates. The best RLCS solution $s_{best}$ is derived when $V_{ext}$ becomes empty by extracting the corresponding complete solutions from the nodes in the current beam.

\begin{algorithm}[!t]
	\caption{BS for the RLCS problem.}\label{alg:bs_rlcs}
	\begin{algorithmic}[1]
		\Statex \textbf{Input}: Instance $I$ of dimension $m\times k$; heuristic function $h$; beam width $\beta$
		\Statex \textbf{Output}: a (heuristic) solution $s$
		\State $r \gets ((1, \ldots, 1), (1, \ldots, 1), 0)$
		\State $s_{best} \gets \varepsilon$
		\State $B \gets \{r\}$
		\While{$B \neq \emptyset$}
		\State $V_{ext} \gets \emptyset$
		\For{$v \in B$}
		\State $children \gets \texttt{expand\_node}(v)$
		\State $V_{ext} \gets V_{ext} \cup children$
		\EndFor
		\State $V_{ext} \gets \texttt{sort\_nodes}(V_{ext}, h) $
		\If{$V_{ext}$ is empty}
		\State $s_{best} \gets$ Derive a partial solution that corresponds to a node in $B$
		\EndIf
		\State $B \gets \texttt{reduce}(V_{ext}, \beta)$
		\EndWhile
		\State \textbf{return} $s_{best}$
	\end{algorithmic}
\end{algorithm}

\subsection{Upper bounds}\label{sec:upper_bounds}

Note that any upper bound for an LCS problem instance is also an upper bound for the corresponding RLCS problem instance obtained by adding a set $R$ of restricted strings to the LCS instances' input strings. The upper bound we used within A$^*$ search and BS is the minimum of two known LCS upper bounds, denoted as $\text{UB}_1$ and $\text{UB}_2$. $\text{UB}_1$ is hereby adapted to the RLCS problem as explained further down. Detailed information on these bounds is given in~\cite{blum2009beam,wang2010fast}. For the sake of completeness, we provide some essential information on these two upper bounds here.  \\

The upper bound $\text{UB}_1$ of an LCS problem instance $(S, \Sigma)$ determines for every letter from $\Sigma$ an upper bound on the number of times this letter potentially may appear in an optimal solution, and then determines the sum of these values across all letters. The number of times letter $a \in \Sigma$ may appear in an optimal solution is evidently bounded by $\min_{i=1,\ldots, m}\{ |s_i|_a \mid s_i \in S\}$. By summing over all $a\in \Sigma $, we get a valid  upper bound 

\begin{equation}
	\text{UB}_1(S)= \sum_{a \in \Sigma}\min_{i=1,\ldots, m} |s_i[p^{L, v}_i, |s_i|]|_a
\end{equation}

Note that for the definition of the evaluation function $h()$ in A$^*$ search and BS, this upper bound is obviously applied to the remaining sub-instance $(\{s_i[p^{L, v}_i, |s_i|] \mid i =1 \ldots,m\}, \Sigma)$ defined by the state graph node $v$ to be evaluated. In other words, $h(v) := \text{UB}_1(\{s_i[p^{L, v}_i, |s_i|] \mid i =1 \ldots,m\})$. In an abuse of notation, we will also use notation $\text{UB}_1(v)$. Moreover, this bound can be further tightened. Namely, when node $v$ induces a complete solution, this bound is simply set to 0, i.e., the adapted upper bound is  
$$\overline{\text{UB}}_1 (v)  = 
\begin{cases}
	\text{UB}_1(v), \text{ if } v \text{ is not complete}, \\ 
	0, \text{ otherwise}
\end{cases}
$$
Thus, for each child node $v'$ of a node $v$, we must check in advance (look ahead) to see whether or not it induces a complete solution. Depending on the outcome, for example, in the  A$^*$ search,  we set $f(v'):=u^{v'}$ when $v'$ is complete. This ensures that the first complete solution found by A$^*$ search is indeed optimal. Note that this was not the case in our conference paper~\cite{djukanovic_rlcs_search_2024}, where we had to keep track of the progress of the best solution before cutting off equally good or proven sub-optimal nodes to terminate the algorithm with completion. While this did not cause a huge bottleneck in the aforementioned paper due to the size of problem instances, which were all small to middle-sized, this simple bound tightening helped save a considerable amount of time for the A$^*$ search when applied on much larger instances as introduced in this paper. \\

The second upper bound $\text{UB}_2$ utimizes DP to determine the optimal LCS solution of two reversed input strings. In essence, DP is applied to each pair $(s_1^{rev}, s_2^{rev}), \ldots, (s_{m-1}^{rev}, s_m^{rev})$, generating the DP (scoring) matrices $M_i, i=1, \ldots, m-1$. Note that at a position $M_i[x, y]$ the length of the LCS solution between $s_i[|s_i| - x + 1, |s_i|]$ and $s_{i+1}[|s_{i+1}| -y+1, |s_{i+1}|]$ is kept in this way, $1 \leq x \leq |s_i|, 1\leq y \leq |s_{i+1}|$, with $M_i[0, y]=M_i[x, 0]=0$, for each $x, y \geq 0$. Thus, when applying this upper bound to the remaining sub-instance regarding a state graph node $v$, it can be expressed as follows: 
\begin{equation}
	\text{UB}_2(v) = \min_{i=1, \ldots, m-1}M_i\left[ |s_i| - p^{L,v}_i + 1, |s_{i+1}| - p^{L,v}_{i+1}+1 \right]
\end{equation}
These $(m-1)$ matrices are preprocessed before executing the algorithm. The preprocessing step is done in $O(n^2\cdot m)$ time.  Finally, for each node $v$, the combined upper bound is given by 
\begin{equation}
	\text{UB}(v)= \min\{\overline{\text{UB}}_1(v), \text{UB}_2(v)\}
\end{equation}
One can easily prove that the upper bound \textsc{UB} is monotonic and thus admissible, i.e., it never underestimates the length of the optimal path from node $v$ to a goal node.  

\subsection{A Probability-based Heuristic Function for RLCS } \label{sec:prob-based-guidance}

This section is devoted to deriving a probability-based search guidance to guide the search. It is based on a probabilistic model constructed assuming input strings are generated uniformly at random. The probabilistic model for the classical LCS problem has been proposed in~\cite{mousavi2012improved}. After improving this model, we utilize it for the RLCS problem, additionally supported by a tie-breaking mechanism.  \\

Let us assume that all strings in $S$ are independently and randomly generated. Given an arbitrary string $s$, let $E_i$ denote the event that $s$ is a subsequence of $s_i$, that is, $s \prec s_i$. Let us denote the probability that the event $E_i$ is realized by $Pr(E_i), i=1,\ldots, n$. In that way, one obtains
\begin{align*}
	&Pr(s \prec S)  = Pr( s \prec s_1 \wedge \ldots \wedge s \prec s_m)  =Pr(\cap_{i=1}^m E_i)\\
	&= \prod_{i=1}^m Pr(E_i) = \prod_{i=1}^m Pr(s \prec s_i) = \prod_{i=1}^m P(|s|, |s_i|),
\end{align*}
where $P$ is a matrix that, at position $(i,j)$, contains the probability that an arbitrary string of length $i$ is a subsequence of a random string of length $j$.  Note that we assume mutual in-dependency between all strings from $S$ here. Note that the DP approach for determining entries of this matrix $P$  is derived in~\cite{mousavi2012improved} and is given by the following recurrence relation:

$$P(i, j) =
\frac{1}{|\Sigma|} P(i-1,j-1) + \frac{|\Sigma|-1}{|\Sigma|}P(i-1, j), \quad \text{for } 0 \leq i,j \leq n,$$

where the initial values are set to $P(i, j)=0$ for $i > j $ and $P(0, j)=1$ for $j\geq 0$.  Matrix $P$ can be pre-processed in $O(n^2)$ time. \\

Now, this result can be seen as a heuristic guidance regarding any node $v=(p^v, l^v, u^v)$ and its associated sub-problem in the following way:
\begin{equation}
	H_{RLCS}(v) = \prod_{i=1}^m P(k, |s_i| - p^v_i+1) 
\end{equation}

where $k$ represents a strategic parameter, chosen heuristically at each level of BS. In~\cite{mousavi2012improved}, 
the following formula is applied to calculate the value of $k$
\begin{equation}\label{eq:mousavli-k-determination}
	k = \max \left \{1,  \left \lceil \min_{v \in   V_{ext} }\frac{|s_i| - p^v_i +1 }{|\Sigma|} \right \rceil \right \} 
\end{equation}

As stated in the original paper~\cite{mousavi2012improved}, the chosen value of $k$ may actually be improved. And in fact, we detected the following issue with Eq.~(\ref{eq:mousavli-k-determination}). Namely, utilizing all nodes from $V_{ext}$ for calculating an appropriate $k$-value, as done in Eq.~(\ref{eq:mousavli-k-determination}),  may be inappropriate and far away from reality for many nodes, as the optimal length from a node at the considered level to goal nodes is usually much higher than that given by Eq.~(\ref{eq:mousavli-k-determination}). Thus, $k$ may be underestimated in this way. Note that the $k$-value refers to an estimated number of times partial solutions of \emph{each} node in $V_{ext}$ can be extended. By utilizing unreasonably small $k$ value, the importance of, in reality, more promising nodes can be marginalized and make them close to those that are less promising. This issue frequently affects the search at deeper levels of the beam search where nodes are closer to goal nodes, and the value of $k$ often reduces to the minimum value of 1 too quickly according to Eq.~(\ref{eq:mousavli-k-determination}). To fix this issue, we employ the following methodology. \\

We determine a more suitable value of $k$ on the basis of a subset $V'_{ext} \subseteq V_{ext}$ of predefined promising nodes.  To determine $V'_{ext}$  at each level,   all nodes are sorted according to their upper bound (UB) value in a decreasing order. A tie-braking mechanism is employed for doing so. This is done by leveraging the so-called $R_{\min}$ score defined by
\begin{equation}
	R_{\min}(v) = \min \{ |r_i|- l^v + 1 \mid i=1, \ldots, k  \}
\end{equation} 
where larger values are favored. Then, a \textit{percent\_extensions} (parameter of the algorithm) of the leading nodes from $V_{ext}$ are pursued to $V'_{ext}$. Afterwards, Eq.~(\ref{eq:mousavli-k-determination}) is applied on the basis of $V'_{ext}$ (instead of $V_{ext}$) for determining the value of $k$. In case two nodes with the same $H_{RLCS}$-value exist, again, the one with a larger $R_{\min}$-score is given preference.

\section{Learning Beam Search Approach}\label{sec:learning-bs}

Finally, after describing an upper bound and a heuristic guiding function for the RLCS problem, we also introduce a heuristic function obtained by learning. The neural network model proposed for this purpose is trained offline, utilizing a set of training and validation problem instances. For providing heuristic guidance regarding a state graph node $v$, the neural network receives a set of (numerical) node features to be evaluated and provides the heuristic guidance value as output.

In this context, note that, recently, several BS approaches based on learned heuristic functions have been proposed, see~\cite{huber2021learning,huber2022relative,ettrich2023policy}. These methods from the literature are developed in the context of the LCS problem and the Constrained Longest Common Subsequence Problem (CLCS), yet another practically motivated LCS problem variant. Similarly to our work, \cite{huber2021learning,huber2022relative} build a neural network to predict the heuristic value of the nodes at each level of BS, while~\cite{ettrich2023policy} learn a policy to select the most promising ones. On top of being applied to a different LCS problem variant, our proposed approach differs from these frameworks in the training process. The mentioned methods from the literature utilize ideas from Reinforcement Learning (RL) to train the neural network in contrast to an evolutionary algorithm (EA) used here.  In the future, it makes sense to compare our method to these three approaches from the literature by solving the same set of (combinatorial) optimization problems under the same conditions.

\subsection{Features}

Two types of features are used as input of the neural network for every state graph node $v$ to be evaluated: (i) features related to the $v$, which try to capture its characteristics, and (ii) general features related to the problem instance under consideration, such as the number of input strings, the number of restricted strings, and the alphabet size. 

\subsubsection{Node features}

Remember that a node $v$ is stored as a tuple $(p^{L,v}, l^v, u^v)$ in the state graph. Vector $p^{L,v}$ keeps track of the input strings' relevant parts (suffixes) available for further extension of the partial solution represented by the $v$. Vector $l^v$ keeps track of the suffixes of the restricted strings that are subsequences of this same partial solution, and finally, $u^v$ is the node's partial solution length. These three values are used to define the node features. \\ 

Note that the length of vectors $p^{L,v}$ and $l^v$ depends on the amount of input and restricted strings, respectively. On top of this, the scale of their values depends on the length of these strings. To keep the number of features and their scale comparable throughout instance sizes, their information is summarized in the following manner. First of all, both vectors are normalized regarding string length in the following way:

\[ \Tilde{p}^{L,v}_i = \frac{p^{L,v}_i}{|s_i|} \ \text{for} \ i\in\{1,\dots, m\}  \ \text{and} \
\Tilde{l}^{v}_j = \frac{l^v_j}{|r_j|} \ \text{for} \ j\in\{1,\dots, k\}
\]

The maximum, minimum, average, and standard sample deviation of the resulting standardized vectors are then used as the corresponding features, making the number of features independent of the instance size:
\[\Big(\max\big(\Tilde{p}^{L,v}\big), \min\big(\Tilde{p}^{L,v}\big), \text{avg}\big(\Tilde{p}^{L,v}\big), \text{sd}\big(\Tilde{p}^{L,v}\big), \max\big(\Tilde{l}^{v}\big), \min\big(\Tilde{l}^{v}\big), \text{avg}\big(\Tilde{l}^{v}\big), \text{sd}\big(\Tilde{l}^{v}\big)\Big)\]

Hence, this results in a total of eight node features to be utilized as input to the neural network.

\subsubsection{Instance features}

To provide information about the specific problem instance, the following features are included: (i) alphabet size ($|\Sigma|$), (ii) number of input strings (m), and (iii) number of restricted strings ($k$). Moreover, for the \textsc{Random} benchmark set, the length of the input strings ($n$) and of the restricted strings ($r_0$) are also used as two additional features, as in these problem instances all input strings and all restricted strings have the same length, respectively.

Therefore, in total 13 and 11 features are extracted when tackling a problem instance from benchmark sets \textsc{Random} and \textsc{Abstract}, respectively. Finally, note that before the neural network takes the features as input, these are normalized to have unit-mean and zero-variance.

\begin{figure}
	\centering
	\includegraphics[width=300pt,height=170pt]{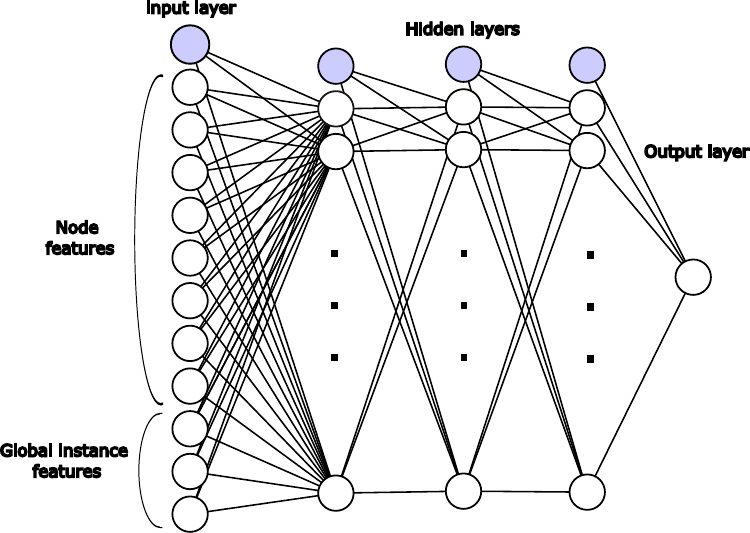}
	\caption{A graphical representation of the feed-forward neural network employed for benchmark set \textsc{Abstract}. The lines from the top node in each layer represent the biases. Remember that two extra instance features are considered in the context of benchmark set \textsc{Random}, as opposed to the benchmark set \textsc{Abstract}.}
	\label{fig:nnet}
\end{figure}

After conducting several preliminary experiments, we decided to use a feed-forward neural network that consists of three hidden layers; the first two hidden layers comprise 10 nodes each, and the last hidden layer comprises 5 nodes. All three layers employ the \texttt{sigmoid} activation function. Figure~\ref{fig:nnet} illustrates the structure of the considered neural network.

\subsection{Neural Network Training}

As mentioned earlier, we trained the neural network on full-size instances using an EA. Hereby, each individual represents a set of weights for the neural network, and the EAs' population is evolved until overfitting is detected. Contrary to the traditional supervised approach, this method does not require having examples for training. This is particularly desirable in our context due to the difficulty of obtaining optimal node-heuristic value pairs to be used as training examples, especially for large-sized instances. \\

The training process employs two sets of problem instances, which we henceforth call \emph{training} and \emph{validation} instances, denoted by $T_{\text{inst}}$  and $\emph{VL}_{\text{inst}}$, respectively. The evaluation of an individual---that is, the evaluation of a set of weights---works as follows. Firstly, the neural network is equipped with the weights of the individual. Afterward, the BS guided by the neural network is applied to every instance from the training set $T_{\text{inst}}$. The individual's fitness is then set to the average length of the obtained solutions, which is referred to as the training value. The overfitting issue is checked once a new best individual is found, representing an individual with a new best training value. This step consists of again executing BS guided by the neural network, but this time on the validation instances from $\emph{VL}_{\text{inst}}$. The average length of the solutions obtained refers to the validation value of the individual. Validation values are used to decide when to stop the training process. In particular, we decided to train in an early stopping fashion, terminating the training process whenever the validation value decreases.

\begin{algorithm}[!t]
	\caption{The \textsc{Brkga} used for training}
	\begin{algorithmic}[1]
		\Statex \textbf{Input}: Values for parameters
		$p_{\text{size}},p_\mathrm{e},p_\mathrm{m},\text{ and }\rho_\mathrm{e}$. Moreover, $T_{\text{inst}}$: training instances, $\emph{VL}_{\text{inst}}$: validation instances
		\Statex \textbf{Output}: A weight value setting for the neural network
		\State \textbf{for} $j=1,\dots, p_{\text{size}}$ \textbf{do} 
		\State \ \ \ \ \ $P \gets P \cup \{\textsf{generate\_random\_individual}()\}$
		\State \textbf{end for}
		\State $ \emph{best\_validation\_value} \gets$   \textsf{evaluate}($P, T_{\text{inst}}, V_{\text{inst}}$)
		\State \emph{overfit} $\gets false$
		\While{termination conditions not met $\wedge$  $!$\emph{overfit}}
		\State $P_\mathrm{e} \gets$ the best $p_\mathrm{e}$ individuals from $P$
		\State $P' \gets P_\mathrm{e}$ 
		\For{$j= 1,\dots,p_\mathrm{m}$}
		\State $\mathbf{v} \gets \textsf{generate\_random\_individual}()$
		\State $P' \gets P' \cup  \{ \mathbf{v} \}$
		\EndFor  
		\For{$j=1,\dots,p_{\text{size}}-p_\mathrm{e}-p_\mathrm{m}$}
		\State select $\mathbf{v}^1 \in P_{\mathrm{e}}$ and $\mathbf{v}^2 \in P\setminus P_{\mathrm{e}}$ randomly
		\State \textbf{for} $i = 1,\dots, m $ \textbf{do}
		\State \ \ \ \ \ \textbf{with} probability $\rho_\mathrm{e}$ \textbf{do}
		\State \ \ \ \ \ \ \ \ \ \ $v_i\gets v^1_i$
		\State \ \ \ \ \ \textbf{otherwise}
		\State \ \ \ \ \ \ \ \ \ \ $v_i\gets v^2_i$
		\State \textbf{end for}
		\State $P' \gets P' \cup  \{\mathbf{v}\}$ 
		\EndFor
		\State $P \gets P'$ 
		\State $\emph{validation\_value} \gets \textsf{evaluate}(P, T_{\text{inst}}, V_{\text{inst}})$ \hspace{0.4cm} \# sort $P$ 
		\If{\emph{validation\_value} not \emph{null}} 
		\If{$\emph{validation\_value} < \emph{best\_validation\_value}$}
		\State \emph{overfit} $\gets$ \emph{true}
		\Else
		\State $ \emph{best\_validation\_value} \gets \emph{validation\_value} $ 
		\EndIf
		\EndIf
		\EndWhile
		\State \textbf{Return: } $P$[0]
	\end{algorithmic}
	\label{alg:brkga_general_structure}
\end{algorithm}

The particular EA used for training is a so-called \emph{Biased Random Key Genetic Algorithm} (BRKGA), initially introduced in~\cite{gonccalves2011biased} as a variant of the classical GA~\cite{goldberg1989genetic,holland1975genetic}. GAs work on a population of individuals, each one representing a solution to the optimization problem at hand. As mentioned above, this is done in our case through a set of weights for the neural network. The population evolves throughout several iterations, called generations. In each generation, a new population of individuals is constructed from the current one, employing nature-inspired operators such as mating and mutation. During execution, individuals with higher fitness are preferred to increase the overall population fitness over time. 

In a classical BRKGA, an individual $\mathbf{v} = (v_1,\ldots,v_r)$ is a vector of $r$ real numbers, generally from the interval $[0,1]$. In our case, we considered individuals as vectors of numbers from $[-1,1]$ so that weights are not restricted to non-negative values. Moreover, $r$ is set to the number of weights of the neural network. A population $P$ of $p_{\text{size}}$ individuals is maintained. The population is initialized with random individuals, that is, with random values from $[-1,1]$. Subsequently, the population is evaluated by computing the training value of each individual, done by executing BS on the set of training instances from $T_{\text{inst}}$. The population is then split into two parts: (1) the \emph{elite} population $P_\mathrm{e} \subset P$ that consists of the best $p_\mathrm{e}$ individuals of $P$ and (2) the \emph{non-elite} population, consisting of the remaining ones. The number $p_\mathrm{e} < p_{\text{size}} - p_\mathrm{e}$ is a parameter called the number of elites. Another parameter $p_\mathrm{m} < p_{\text{size}}-p_\mathrm{e}$, called the number of mutants, is then used to generate the next population of individuals. This is done by passing the elite population to the next generation along with $p_\mathrm{m}$ mutant individuals which, like the initial population, are constructed randomly. The remaining $p_{\text{size}}-p_\mathrm{e}-p_\mathrm{m}$ individuals are introduced through the process of \emph{mating}. For each individual, two parents are selected randomly, one from the elite population and one from the non-elite population. Then, the $i$-th vector position of the offspring individual is set to the $i$-th vector position of one of the two selected parents, choosing between the two depending on a parameter $\rho_\mathrm{e} \in (0.5, 1]$, called the elite inheritance probability. Note that the this way of working is absolutely standard for any BRKGA, as originally described in~\cite{gonccalves2011biased}. \\

Algorithm~\ref{alg:brkga_general_structure} shows a pseudo-code for the BRKGA used for training. Note that function $\textsf{evaluate}(P, T_{\text{inst}}, \emph{VL}_{\text{inst}})$ computes the training value of the individuals of the population. It additionally returns the validation value (\emph{validation\_value}) in case a new best individual is found; note that the \emph{validation\_value} of a specific iteration equals to  \emph{null} if a new best individual has not been found for that iteration. Moreover, remember that the training is terminated once the validation value decreases. 


\section{Experimental Evaluation}\label{sec:experimental_evaluation}  

This section provides a comprehensive experimental comparison between the following four competitors: 
\begin{itemize}
	\item The exact A$^*$ search, described in Section~\ref{sec:a_star};
	\item The three versions of the BS algorithm: 
	\begin{itemize}
		\item the BS guided by UB (labeled by BS-ub);
		\item the BS guided by the probability-based guidance, drafted in Section~\ref{sec:prob-based-guidance} (labeled by BS-prob);
		\item the learning BS, as drafted in Section~\ref{sec:learning-bs}  (labelled by LBS). 
	\end{itemize}
\end{itemize}
Note that all other (exact) approaches from the literature for the RLCS problem were shown to be inferior to BS-ub and A$^*$ search in~\cite{djukanovic_rlcs_search_2024}, where the evaluation is performed for much smaller instances introduced earlier in the literature. The same is concluded with our preliminary experimental evaluation of the newly introduced datasets. Hence, these approaches are omitted from further analysis and are therefore not presented here. All experiments were conducted in single-threaded mode on an Intel Xeon E5-2640 with 2.40GHz and 16 GB of memory for the heuristic approaches and 32 GB for the A$^*$ search approach.  All instances with corresponding binaries of our LBS approach are available at the Git repository  accessible through the link \url{https://github.com/markodjukanovic90/RLCS-LBS}.

\subsection{Problem instances}

The experimental evaluation employs two newly generated benchmark sets. 
\begin{itemize}
	\item Benchmark set \textsc{Random} comprises instances built from randomly generated strings. 5 random instances were generated for each combination of $n \in \{200, 500, 1000\}$, $m \in \{3, 5, 10\},  p \in \{3, 5, 10\}$, $|p_0| \in \{ 1\%, 2\%, 5\% \}$ (of length of input strings), and $|\Sigma| \in \{4, 20\}$. Overall, $3 \cdot 3 \cdot 3 \cdot 3 \cdot 2 \cdot 5 =810$ random RLCS instances are included in this dataset.  
	\item Benchmark set \textsc{Abstract}  comprises 298 instances where the core input strings are the input strings from the \textsc{Abstract} dataset primarily used for the LCS problem~\cite{nikolic2021solving}.  These input strings are characterized by close-to-polynomial distributions of different letters from the English alphabet. The input strings originate from abstracts of real scientific papers written in English. Pattern strings are added to these input strings to create RLCS problem instances as follows. In particular, the 60  most frequently occurring words in the research corpus were taken as pattern strings, see~\cite{coxhead2002academic,mozaffari2014academic,coxhead2000new}. Our intention in making use of these patter strings was to verify the overall effect of commonly used words on the conclusion about plagiarism/similarity between abstracts. 
	
	The instances of this dataset are split into two categories: POS and NEG. The first group consists of instances for which it is known that there is a positive correlation of similarity between abstracts. In contrast, for the latter (NEG) group, an opposite conclusion is known to be valid. 
	In detail,  by using \emph{tf-idf} statistics with cosine similarity, the algorithm from~\cite{magara2018comparative}  identified similar papers from a large paper collection. After that, the similarity between the abstracts of papers proposed by that algorithm was manually checked and tagged by an expert as either similar (positive) or dissimilar (negative). The results of this research can be found at \url{https://cwi.ugent.be/respapersim}. 
\end{itemize}

\subsection{Parameter tuning} 

\begin{figure}[!t]
	\centering
	\begin{subfigure}[b]{0.45\textwidth}
		\centering
		\includegraphics[width=\textwidth]{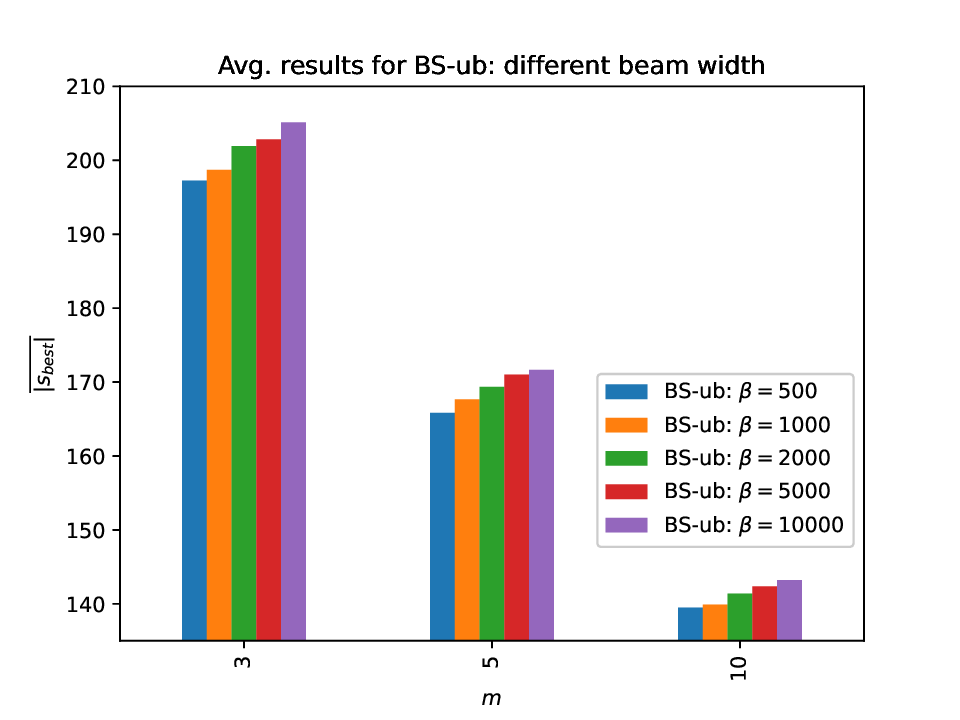}
		\caption{BS-ub: solution quality}
		\label{fig:bs-ub-random-tuned}
	\end{subfigure}
	\hfill
	\begin{subfigure}[b]{0.45\textwidth}
		\centering
		\includegraphics[width=\textwidth]{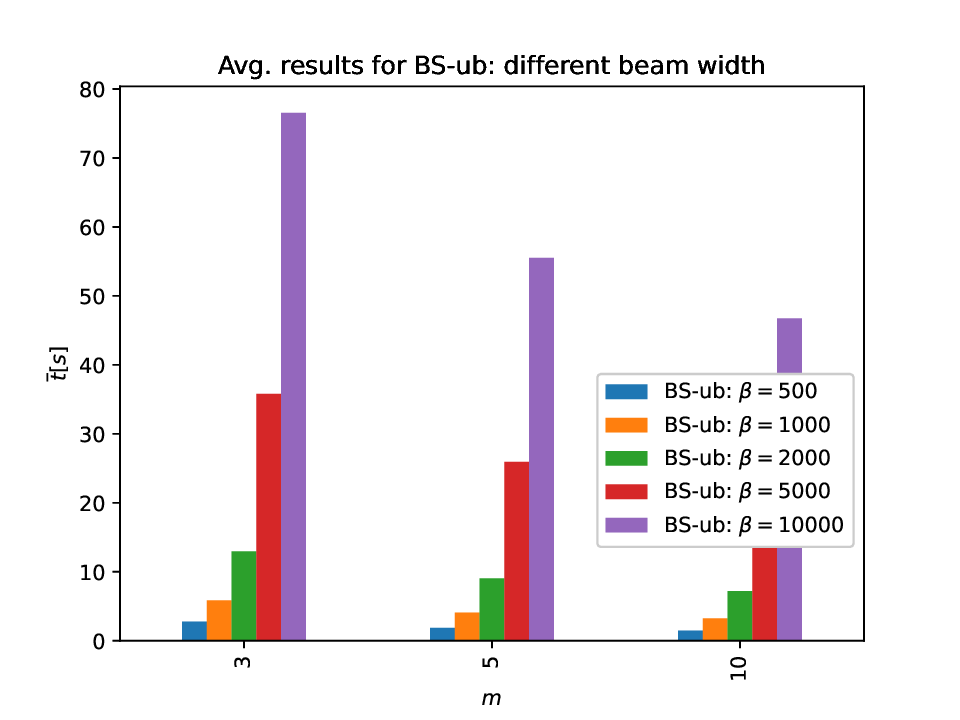}
		\caption{BS-ub: running time}
		\label{fig:bs-ub-random-tuned-runtime}
	\end{subfigure}
	
		\begin{subfigure}[b]{0.45\textwidth}
			\centering
			\includegraphics[width=\textwidth]{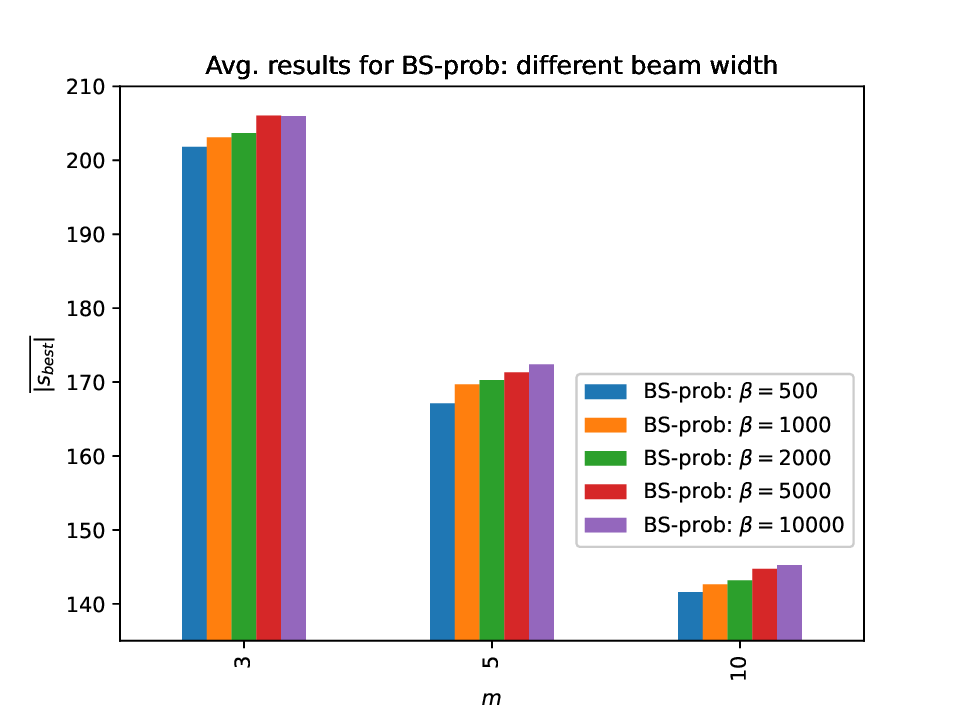}
			\caption{BS-prob: solution quality}
			\label{fig:bs-prob-random-tuned}
		\end{subfigure}
		\hfill
		\begin{subfigure}[b]{0.45\textwidth}
			\centering
			\includegraphics[width=\textwidth]{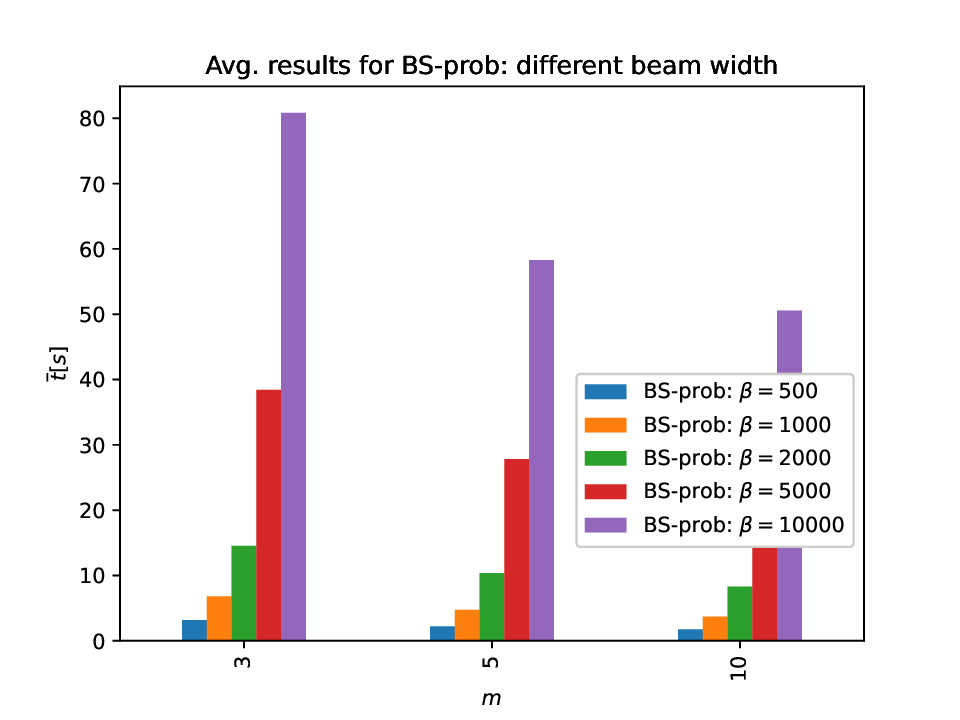}
			\caption{BS-prob: running time}
			\label{fig:bs-prob-random-tuned-runtime}
		\end{subfigure}
		
			\begin{subfigure}[b]{0.45\textwidth}
				\centering
				\includegraphics[width=\textwidth]{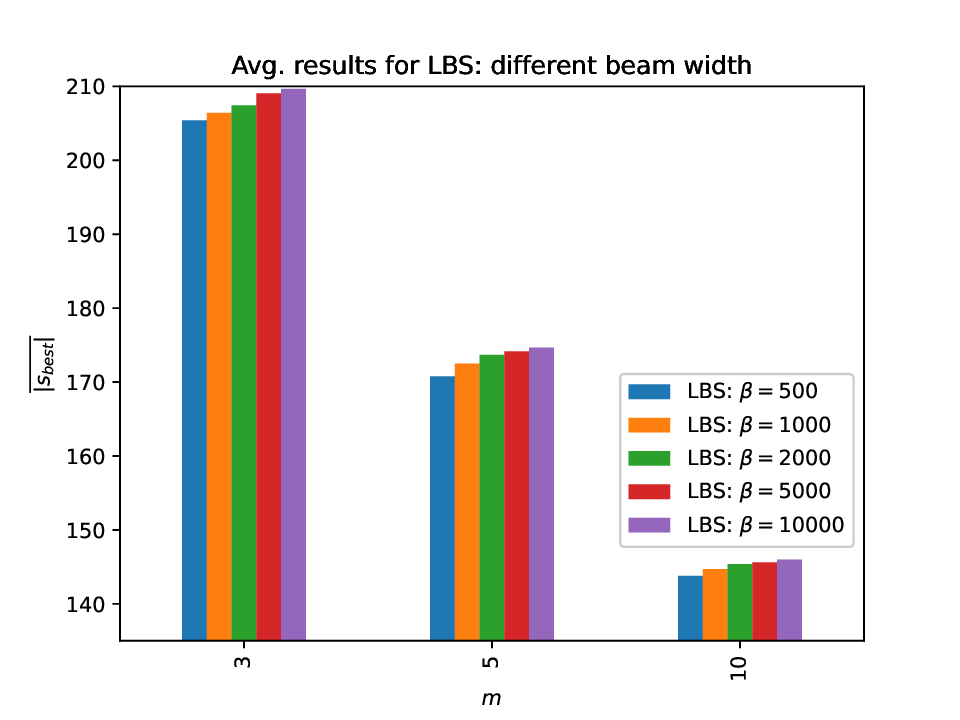}
				\caption{LBS: solution quality}
				\label{fig:lbs-random-tuned}
			\end{subfigure}
			\hfill
			\begin{subfigure}[b]{0.45\textwidth}
				\centering
				\includegraphics[width=\textwidth]{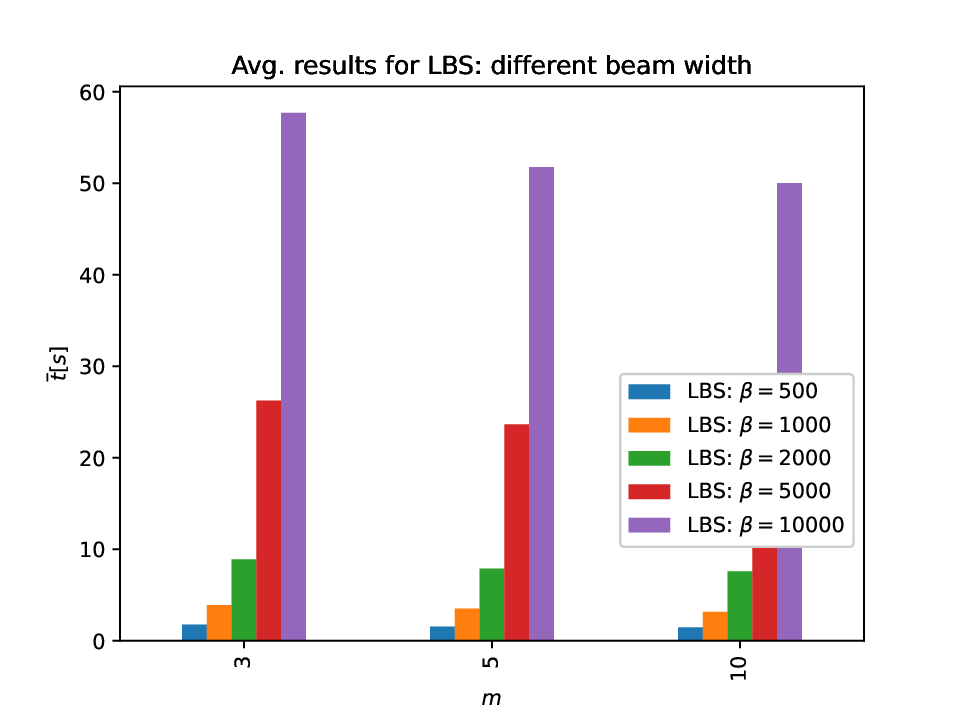}
				\caption{LBS: running time}
				\label{fig:lbs-random-tuned-runtime}
			\end{subfigure}
			\caption{Comparisons of the three BS variants regarding different $\beta$ values on the \textsc{Random} benchmark set}
			\label{fig:lbs-random-report}
		\end{figure}

\begin{figure}[htbp]
			\centering
			\begin{subfigure}[b]{0.45\textwidth}
				\centering
				\includegraphics[width=\textwidth]{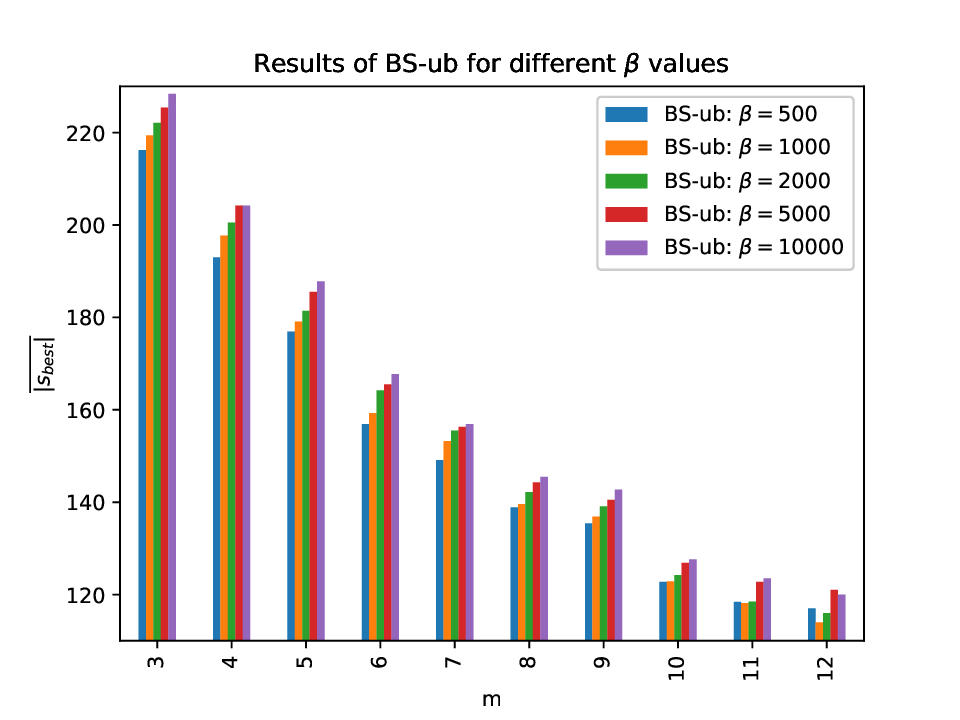}
				\caption{Solution quality (instance type POS)}
				\label{fig:bs-ub-abstract-tuned}
			\end{subfigure}
			\hfill
			\begin{subfigure}[b]{0.45\textwidth}
				\centering
				\includegraphics[width=\textwidth]{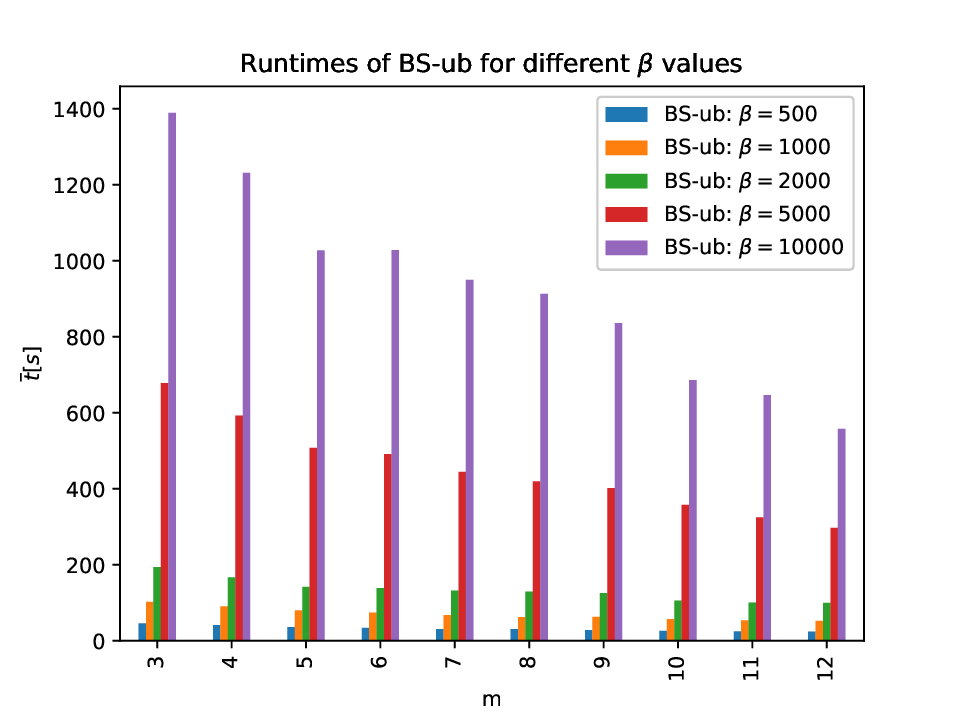}
				\caption{Running time (instance type POS)}
				\label{fig:bs-ub-abstract-pos-tuned-runtime}
			\end{subfigure}
			
			\begin{subfigure}[b]{0.45\textwidth}
				\centering
				\includegraphics[width=\textwidth]{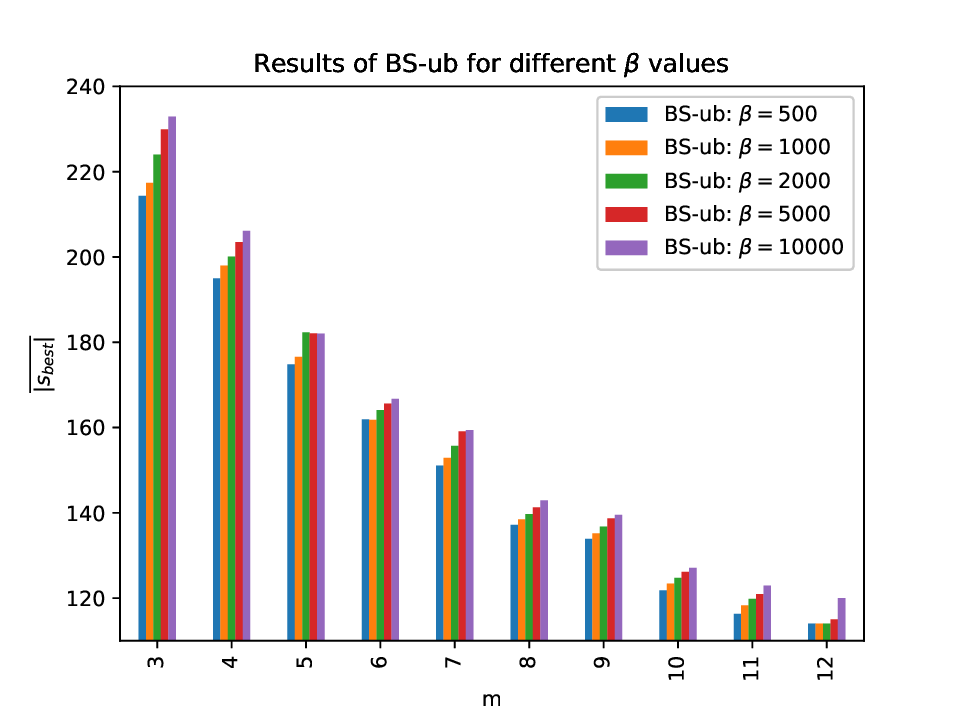}
				\caption{Solution quality (instance type NEG)}
				\label{fig:bs-prob-abstract-tuned}
			\end{subfigure}
			\hfill
			\begin{subfigure}[b]{0.45\textwidth}
				\centering
				\includegraphics[width=\textwidth]{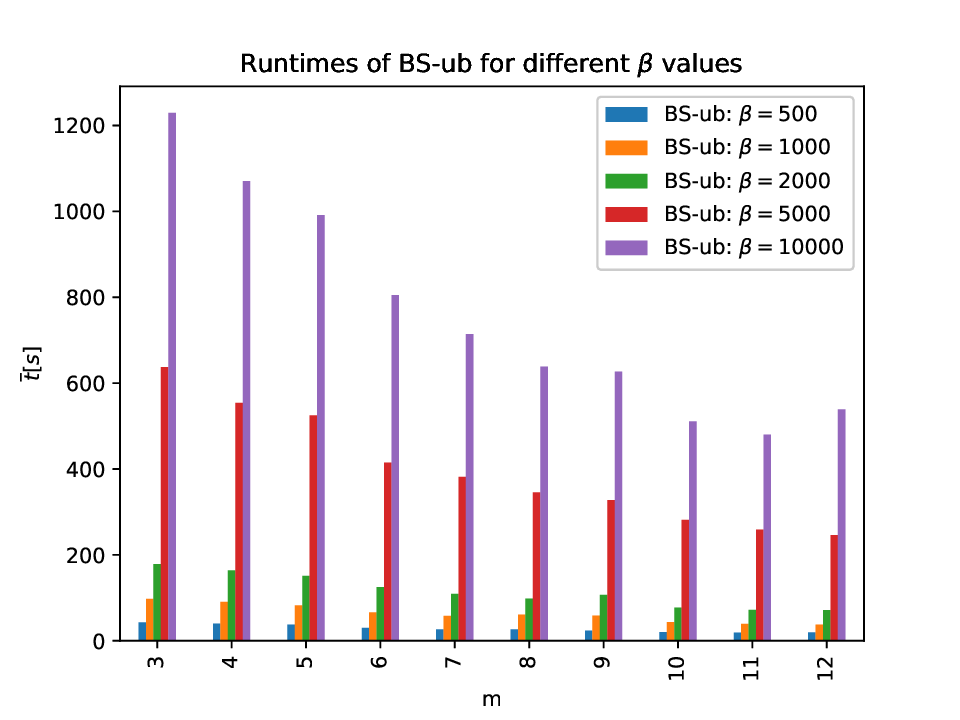}
				\caption{Running time (instance type NEG)}
				\label{fig:bs-ub-abstract-neg-tuned-runtime}
			\end{subfigure}
			\caption{BS-ub: results for different $\beta$ values on benchmark set \textsc{Abstract}}
			\label{fig:bs-ub-abstract-report}
		\end{figure}

\begin{figure}[htbp]
			\centering
			\begin{subfigure}[b]{0.45\textwidth}
				\centering
				\includegraphics[width=\textwidth]{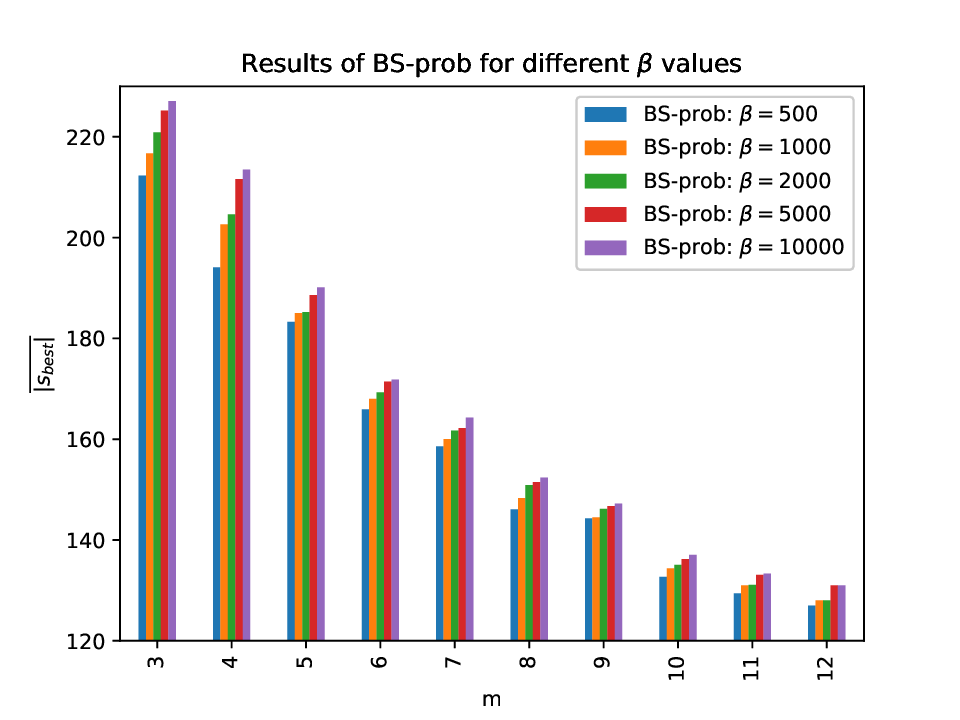}
				\caption{Solution quality (instance type POS)}
				\label{fig:bs-prob-abstract-pos-tuned}
			\end{subfigure}
			\hfill
			\begin{subfigure}[b]{0.45\textwidth}
				\centering
				\includegraphics[width=\textwidth]{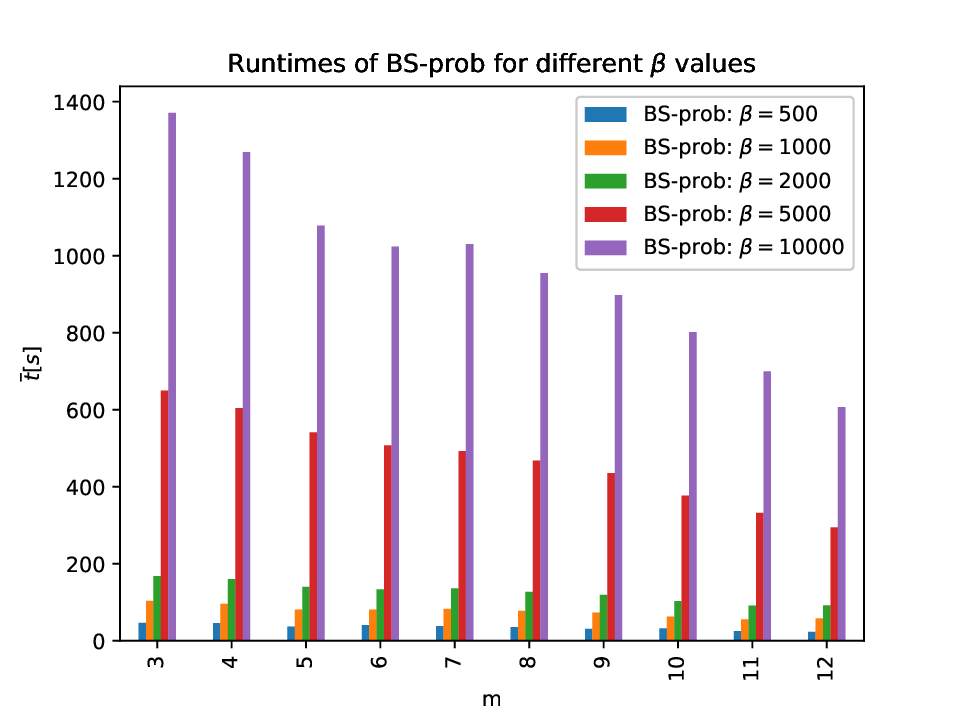}
				\caption{Running time (instance type POS)}
				\label{fig:bs-prob-abstract-pos-tuned-runtime}
			\end{subfigure}
			\label{fig:bs-prob-abstract-pos-report}

			\begin{subfigure}[b]{0.45\textwidth}
				\centering
				\includegraphics[width=\textwidth]{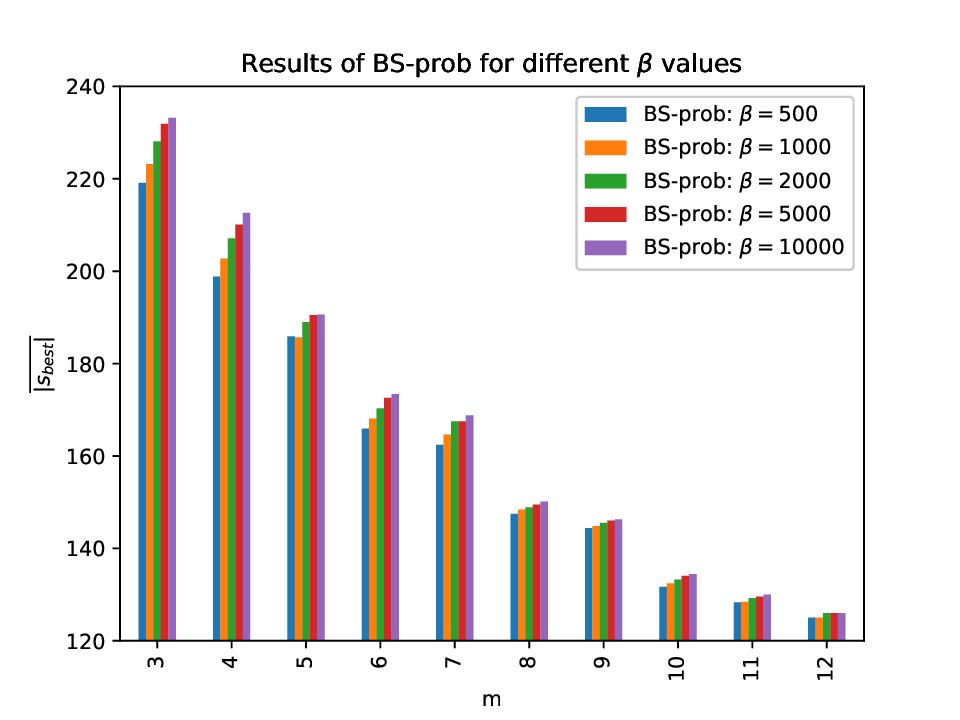}
				\caption{Solution quality (instance type NEG)}
				\label{fig:bs-prob-abstract-neg-tuned}
			\end{subfigure}
			\hfill
			\begin{subfigure}[b]{0.45\textwidth}
				\centering
				\includegraphics[width=\textwidth]{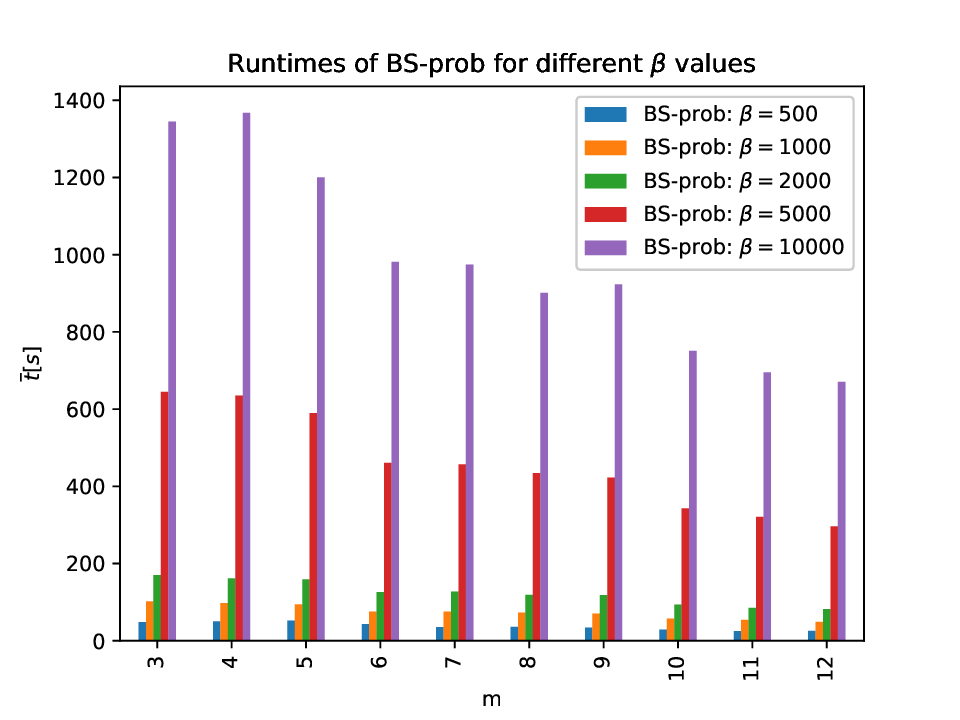}
				\caption{Running time (instance type NEG)}
				\label{fig:bs-prob-abstract-neg-tuned-runtime}
			\end{subfigure}
			\caption{BS-prob: results for different $\beta$ values on benchmark set \textsc{Abstract}}
			\label{fig:bs-prob-abstract}
		\end{figure}

\begin{figure}[htbp]
			\centering
			\begin{subfigure}[b]{0.45\textwidth}
				\centering
				\includegraphics[width=\textwidth]{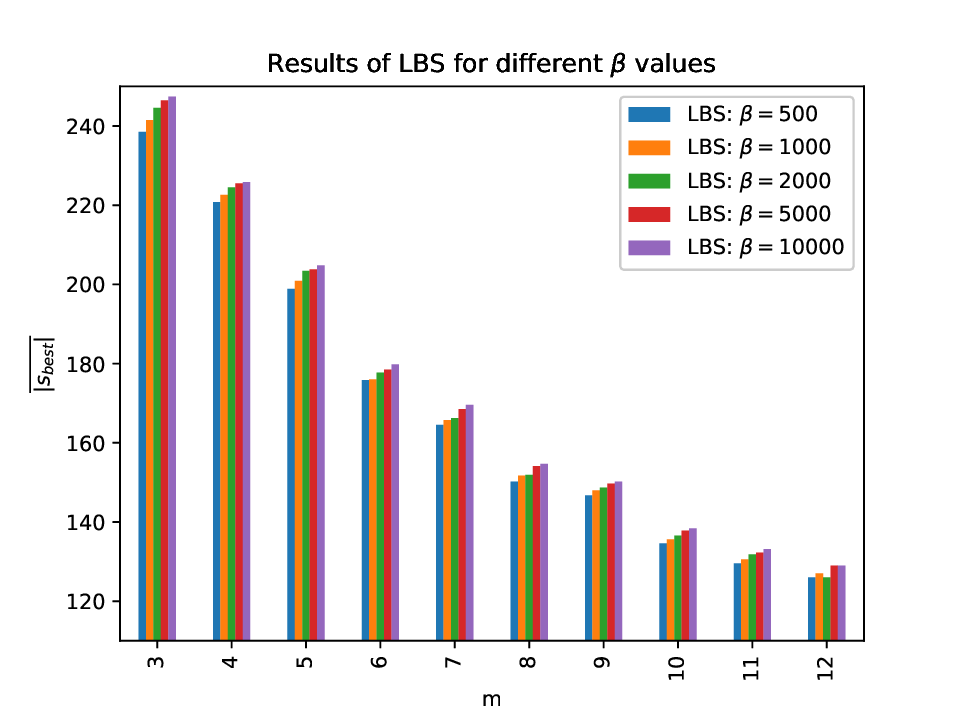}
				\caption{Solution quality (instance type POS)}
				\label{fig:lbs-abstract-pos-tuned}
			\end{subfigure}
			\hfill
			\begin{subfigure}[b]{0.45\textwidth}
				\centering
				\includegraphics[width=\textwidth]{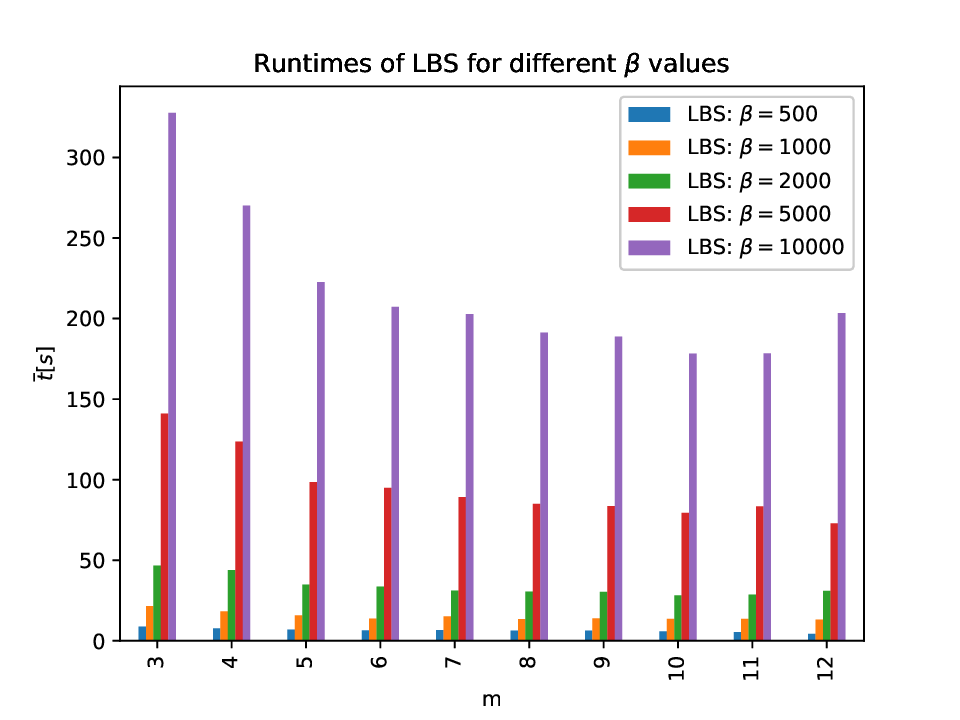}
				\caption{Running time (instance type POS)}
				\label{fig:lbs-abstract-pos-tuned-runtime}
			\end{subfigure}
			\label{fig:lbs-abstract-pos-report}
			
			\begin{subfigure}[b]{0.45\textwidth}
				\centering
				\includegraphics[width=\textwidth]{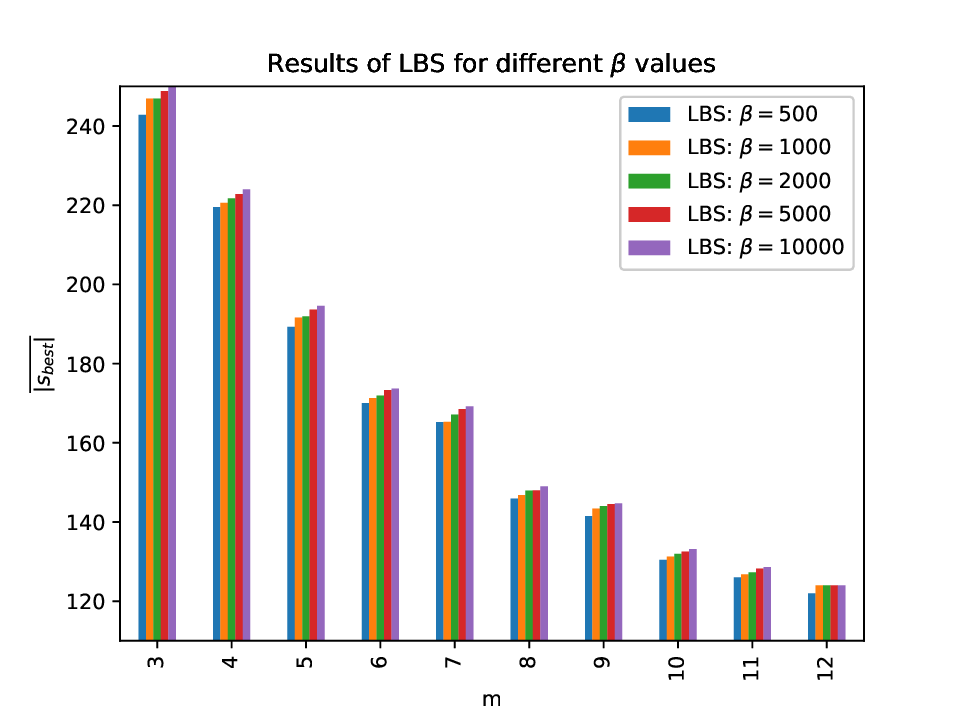}
				\caption{Solution quality (instance type NEG)}
				\label{fig:lbs-abstract-neg-tuned}
			\end{subfigure}
			\hfill
			\begin{subfigure}[b]{0.45\textwidth}
				\centering
				\includegraphics[width=\textwidth]{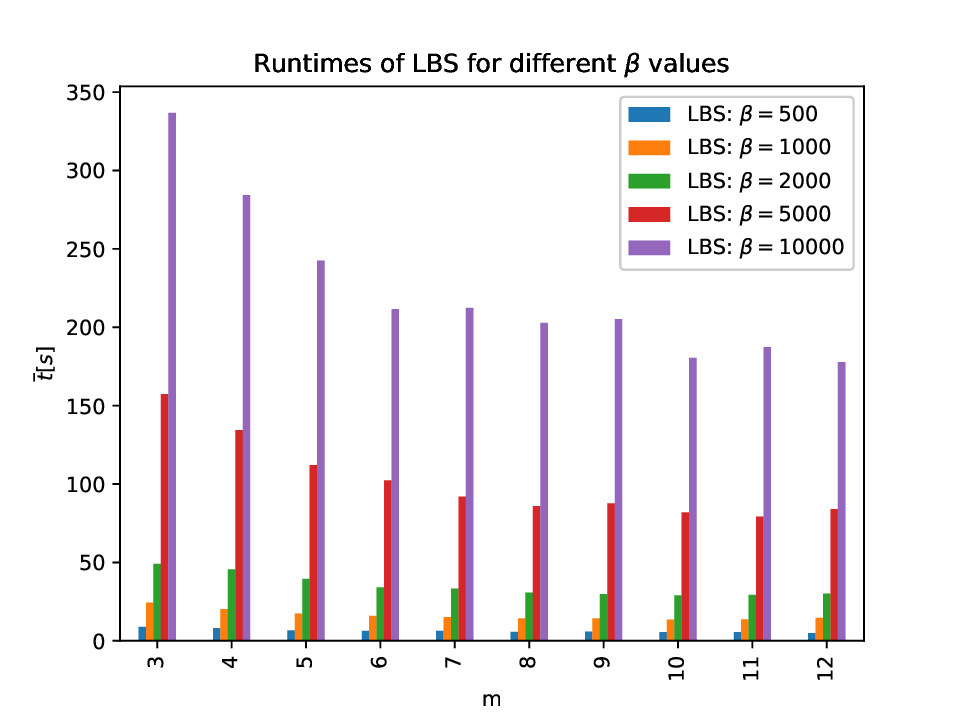}
				\caption{Running time (instance type NEG)}
				\label{fig:lbs-abstract-neg-tuned-runtime}
			\end{subfigure}
			\caption{LBS: results for different $\beta$ values on benchmark set \textsc{Abstract}}
			\label{fig:lbs-abstract}
		\end{figure}

Regarding BS-ub, there is only one parameter to tune ($\beta$). We independently run BS-ub for different $\beta \in \{500, 1000, 2000, 5000, 10000 \}$. Average solution quality over all instances from \textsc{Random} and \textsc{Abstract} benchmark sets per each of the considered $\beta$ values are displayed in Figs.~\ref{fig:bs-ub-random-tuned}, \ref{fig:bs-ub-random-tuned-runtime} and~\ref{fig:bs-ub-abstract-report}, respectively. The best average results are, not surprisingly, reported for the largest value of $\beta$, that is, $\beta=10000$. However, these results are just slightly better than those obtained for $\beta=5000$. However, this improvement comes with 2-3 times higher running times (see Figure~\ref{fig:bs-ub-random-tuned-runtime}, \ref{fig:bs-ub-abstract-pos-tuned-runtime}, and \ref{fig:bs-ub-abstract-neg-tuned-runtime}). Thus, we opt for reporting the results for $\beta=5000$ to aim at high-quality solutions while keeping running times reasonably short. Due to similar reasons, the same is decided for the two other BS derivatives, i.e. BS-prob and LBS, see e.g., Figs.~\ref{fig:bs-prob-random-tuned}--\ref{fig:bs-prob-random-tuned-runtime},  and  Figure~\ref{fig:bs-prob-abstract} in the case of the BS-prob approach, and Figs.~\ref{fig:lbs-random-tuned}--\ref{fig:lbs-random-tuned-runtime} and Figure~\ref{fig:lbs-abstract} in the case of the LBS approach. For both of these BS derivatives, we chose a beam width of $\beta=5000$, which ensured a fair comparison among the approaches in terms of the size of the search space that is examined. We emphasize that after conducting several preliminary experiments,  the \emph{percent\_extensions} parameter is set to $\frac{1}{3} \cdot 100\%$ for the BS-prob approach.

Last but not least, to train the (feed-forward) neural network (NN) of the LBS approach, we employed $\beta=100$ during training for the benchmark set \textsc{Random}. For benchmark set \textsc{Abstract}, we employed $\beta=200$, which seemed to perform better. Remember that this is the value of the beam width used for calculating the \textit{training} and \textit{validation} values during the execution of the training BRKGA. The parameters of this BRKGA were set to the following default values: $p_{\text{size}} = 20$, $p_\text{e} = 1$, $p_\text{m} = 7$ and $\rho = 0.5$. 

\subsection{Numerical results: benchmark set RANDOM} 
\label{subsection:results_random}

Numerical results of the four approaches on the dataset \textsc{Random} are provided in Tables~\ref{tab:random-n-200-m-3}--\ref{tab:random-n-1000-m-10}. These six tables report the results of 6 groups of instances, that is, one for each combination of $n\in \{200, 500, 1000\}$ and $m \in\{2, 10\}$.  The results for instances with $m=5$ are given in Appendix~\ref{sec:appendix-a}.  Each table is divided into five blocks. The first block presents the characteristics of the instance group, including the number of restricted strings ($k$), the length of each restricted string ($|p_0|$), and the alphabet size ($|\Sigma|$). The second block reports on the performance of A$^*$ search, detailing three key metrics: the average solution quality ($\overline{|s_{best}|}$), the average runtime ($\overline{t}[s]$), and the average upper bound ($\overline{ub}$) across five instances for the corresponding group. The final three blocks provide the results for the BS-ub, BS-prob, and LBS algorithms, respectively. Each algorithm is evaluated based on two indicators: average solution quality and average runtime across five instances. For clarity, the best results in each table are highlighted in bold. \\

\noindent The following conclusions are drawn from these results.

\begin{itemize}
	\item  For small-sized instances with $n=200$ and $m=3$, the A$^*$ search successfully finds provenly optimal solutions for 15 out of 18 instance groups, with most of these instances being solved in a relatively short running time. The heuristic approaches demonstrate their effectiveness by finding nearly all these optimal solutions. Moreover, they outperform A$^*$ search on the largest instances with $k=10$ and $|p_0|=10$.  For the instances with $m=5$, A$^*$ search remains a strong performer, optimally solving the instances of 11 instance groups. However, for cases with larger $k$ values and an alphabet size of $|\Sigma|=4$, the problem becomes harder to solve exactly. In these scenarios, the heuristic methods BS-prob and LBS outperform A$^*$, obtaining the best solutions for 14 and 16 instance groups, respectively, while BS-ub follows with successful results in 10 cases. For $m=10$, A$^*$ search can solve only two instance groups optimally. In contrast, LBS and BS-prob are the most effective approaches, delivering the best solutions in 16 and 13 cases, respectively.
	
	\item For instances with $n=500$ and $m=3$, A$^*$ search solves optimally only 3 out of 18 instance groups, primarily due to memory limitations. As expected, the performance of A$^*$ search quickly deteriorates with increasing $n$. Notably, for the largest instances in this group, A$^*$ produces solutions that are far from both the dual bounds and the solutions obtained by the heuristic approaches. LBS outperforms the other methods in these cases, delivering the best performance for instance groups. BS-prob is the second-best approach, achieving the best result in 3 cases. When $m=5$, the performance of A$^*$ search degrades significantly, running out of memory in all cases without reaching the 600-second time limit. In contrast, all three heuristic approaches significantly outperform A$^*$ search, with BS-prob showing the overall best performance in 12 cases, while LBS does so in 10 cases. Both BS-prob and LBS demonstrate similar performance, far exceeding that of BS-ub. For the largest case with $m=10$, the performance of A$^*$ further stagnates, leaving only the heuristic approaches capable of producing high-quality solutions for these medium-sized instances. Both BS-prob and LBS deliver the best solutions for 12 instance groups each. Interestingly, BS-prob tends to perform slightly better than LBS when the number of restricted strings is larger, while LBS shows superiority when $k$ is smaller.

	\item For large-sized instances with $n=1000$ and $m=3$, the most effective approach is LBS, which delivers the best average solution quality in 15 out of 18 cases. BS-prob achieves the overall best result in the remaining 3 cases, while the other two approaches perform significantly worse. For instances with $m=5$, LBS once again outperforms the other methods, providing the best average results in 16 out of 18 cases. BS-prob, in comparison, produces the best results in only 3 cases. LBS remains the clear winner for the largest instances with $m=10$, finding the best results in 13 cases, while BS-prob achieves this in 5 cases. Overall, LBS consistently outperforms the other heuristic approaches, particularly for these largest, and therefore most challenging, instances.
	
\end{itemize}

\begin{table}
\caption{Results for benchmark set \textsc{Random}: $n=200, m=3$}
\label{tab:random-n-200-m-3}
\centering
\begin{tabular}{rrr|lrr|lr|lr|lr}
\toprule
\multicolumn{3}{c}{Instance} &
\multicolumn{3}{c}{A$^*$ search} &
\multicolumn{2}{c}{BS-ub} &
\multicolumn{2}{c}{BS-prob} &
\multicolumn{2}{c}{LBS} \\
 \cmidrule(lrr){1-3}\cmidrule(lrr){4-6}\cmidrule(lr){7-8} \cmidrule(lr){9-10} \cmidrule(lr){11-12} 
 $k$ & $|p_0|$  & $|\Sigma|$ & $\overline{|s_{best}|}$ & $\overline{t}[s]$ & $\overline{ub}$ & $\overline{|s_{best}|}$& $\overline{t}[s]$ & $\overline{|s_{best}|}$ & $\overline{t}[s]$ & $\overline{|s_{best}|}$ & $\overline{t}[s]$\\ \hline
\midrule
3 & 2 & 4 & \textbf{61.2} & 0.44 & 61.20 & \textbf{61.2} & 0.62 & \textbf{61.2} & 0.64 & \textbf{61.2} & 0.46 \\
3 & 2 & 20 & \textbf{44.8} & 7.25 & 44.80 & \textbf{44.8} & 28.38 & \textbf{44.8} & 28.72 & \textbf{44.8} & 4.04 \\
3 & 4 & 4 & \textbf{93.8} & 2.79 & 93.80 & \textbf{93.8} & 2.80 & \textbf{93.8} & 2.88 & \textbf{93.8} & 2.11 \\
3 & 4 & 20 & \textbf{46.0} & 10.53 & 46.00 & \textbf{46.0} & 15.88 & \textbf{46.0} & 16.35 & \textbf{46.0} & 5.93 \\
3 & 10 & 4 & \textbf{106.8} & 32.16 & 106.80 & \textbf{106.8} & 3.39 & 106.60 & 3.32 & \textbf{106.8} & 3.48 \\
3 & 10 & 20 & \textbf{45.8} & 2.99 & 45.80 & \textbf{45.8} & 6.97 & \textbf{45.8} & 6.44 & \textbf{45.8} & 6.59 \\ \hline
5 & 2 & 4 & \textbf{52.8} & 0.29 & 52.80 & \textbf{52.8} & 0.02 & \textbf{52.8} & 0.02 & \textbf{52.8} & 0.02 \\
5 & 2 & 20 & \textbf{43.8} & 12.91 & 43.80 & \textbf{43.8} & 34.45 & \textbf{43.8} & 34.26 & \textbf{43.8} & 4.73 \\
5 & 4 & 4 & \textbf{91.8} & 2.41 & 91.80 & \textbf{91.8} & 2.53 & \textbf{91.8} & 2.65 & \textbf{91.8} & 2.06 \\
5 & 4 & 20 & \textbf{45.2} & 33.52 & 45.20 & \textbf{45.2} & 18.85 & \textbf{45.2} & 18.93 & \textbf{45.2} & 7.52 \\
5 & 10 & 4 & \textbf{106.8} & 227.52 & 106.80 & 104.80 & 3.48 & 105.60 & 3.80 & 106.00 & 3.82 \\
5 & 10 & 20 & \textbf{44.6} & 30.08 & 44.60 & \textbf{44.6} & 9.29 & \textbf{44.6} & 8.55 & \textbf{44.6} & 8.91 \\ \hline
10 & 2 & 4 & \textbf{37.0} & 0.20 & 37.00 & \textbf{37.0} & 0.00 & \textbf{37.0} & 0.00 & \textbf{37.0} & 0.00 \\
10 & 2 & 20 & \textbf{41.0} & 149.93 & 41.00 & \textbf{41.0} & 36.18 & \textbf{41.0} & 35.73 & \textbf{41.0} & 6.18 \\
10 & 4 & 4 & \textbf{87.8} & 1.61 & 87.80 & \textbf{87.8} & 2.43 & \textbf{87.8} & 2.52 & \textbf{87.8} & 1.76 \\
10 & 4 & 20 & 42.80 & 510.59 & 45.80 & \textbf{44.2} & 28.95 & \textbf{44.2} & 30.50 & \textbf{44.2} & 11.33 \\
10 & 10 & 4 & 72.80 & 370.00 & 116.80 & 100.40 & 3.81 & 101.00 & 3.69 & \textbf{101.2} & 3.62 \\
10 & 10 & 20 & \textbf{45.0} & 220.30 & 45.20 & 44.60 & 12.17 & \textbf{45.0} & 11.64 & 44.80 & 12.50 \\ \hline
\bottomrule
\end{tabular}

\end{table} 
\begin{table}
\caption{Results for benchmark set \textsc{Random}: $n=200, m=10$}
\label{tab:random-n-200-m-10}
\centering
\begin{tabular}{rrr|lrr|lr|lr|lr}
\toprule
\multicolumn{3}{c}{Instance} &
\multicolumn{3}{c}{A$^*$ search} &
\multicolumn{2}{c}{BS-ub} &
\multicolumn{2}{c}{BS-prob} &
\multicolumn{2}{c}{LBS} \\
 \cmidrule(lrr){1-3}\cmidrule(lrr){4-6}\cmidrule(lr){7-8} \cmidrule(lr){9-10} \cmidrule(lr){11-12} 
 $k$ & $|p_0|$  & $|\Sigma|$ & $\overline{|s_{best}|}$ & $\overline{t}[s]$ & $\overline{ub}$ & $\overline{|s_{best}|}$& $\overline{t}[s]$ & $\overline{|s_{best}|}$ & $\overline{t}[s]$ & $\overline{|s_{best}|}$ & $\overline{t}[s]$\\ \hline
\midrule
 3 & 2 & 4 & 53.60 & 146.23 & 75.60 & 65.60 & 1.23 & 65.80 & 1.27 & \textbf{66.0} & 1.37 \\
3 & 2 & 20 & 15.40 & 595.00 & 29.20 & \textbf{19.4} & 13.08 & \textbf{19.4} & 15.05 & \textbf{19.4} & 4.58 \\
3 & 4 & 4 & 47.00 & 353.00 & 103.40 & 77.20 & 3.09 & 78.00 & 3.23 & \textbf{78.2} & 3.29 \\
3 & 4 & 20 & 16.60 & 456.00 & 27.20 & 19.40 & 5.96 & \textbf{19.6} & 6.59 & \textbf{19.6} & 4.89 \\
3 & 10 & 4 & 34.40 & 248.00 & 116.60 & 81.20 & 3.49 & 80.80 & 3.72 & \textbf{81.6} & 4.06 \\
3 & 10 & 20 & 16.60 & 371.00 & 25.20 & 19.00 & 4.86 & \textbf{19.2} & 5.46 & \textbf{19.2} & 4.65 \\ \hline
5 & 2 & 4 & \textbf{58.6} & 0.29 & 58.60 & \textbf{58.6} & 0.01 & \textbf{58.6} & 0.01 & \textbf{58.6} & 0.01 \\
5 & 2 & 20 & 14.80 & 595.00 & 29.40 & \textbf{18.6} & 14.20 & \textbf{18.6} & 15.00 & \textbf{18.6} & 4.45 \\
5 & 4 & 4 & 54.80 & 300.24 & 92.60 & 75.20 & 2.73 & 75.00 & 2.85 & \textbf{75.8} & 2.67 \\
5 & 4 & 20 & 15.80 & 456.00 & 29.00 & \textbf{19.2} & 6.19 & \textbf{19.2} & 7.08 & \textbf{19.2} & 5.02 \\
5 & 10 & 4 & 35.80 & 266.00 & 116.20 & 81.40 & 3.93 & \textbf{82.2} & 4.11 & 82.00 & 4.36 \\
5 & 10 & 20 & 16.20 & 354.00 & 28.60 & \textbf{19.2} & 5.31 & \textbf{19.2} & 5.92 & \textbf{19.2} & 5.32 \\ \hline
10 & 2 & 4 & \textbf{36.0} & 0.19 & 36.00 & \textbf{36.0} & 0.00 & \textbf{36.0} & 0.00 & \textbf{36.0} & 0.00 \\
10 & 2 & 20 & 14.40 & 595.00 & 31.00 & 18.40 & 16.89 & \textbf{18.6} & 16.77 & \textbf{18.6} & 4.27 \\
10 & 4 & 4 & 57.40 & 154.71 & 80.80 & 70.40 & 2.37 & \textbf{70.6} & 2.54 & 70.40 & 1.90 \\
10 & 4 & 20 & 15.40 & 414.00 & 30.40 & \textbf{19.4} & 7.01 & \textbf{19.4} & 7.56 & \textbf{19.4} & 5.40 \\
10 & 10 & 4 & 33.40 & 277.00 & 117.20 & 81.20 & 4.10 & 81.80 & 4.23 & \textbf{82.0} & 4.14 \\
10 & 10 & 20 & 15.00 & 307.00 & 30.60 & 19.00 & 6.14 & \textbf{19.2} & 6.66 & \textbf{19.2} & 5.75 \\ \hline
\bottomrule
\end{tabular}
\end{table} 
\begin{table}
\caption{Results for benchmark set \textsc{Random}: $n=500, m=3$}
\label{tab:random-n-500-m-3}
\centering
\begin{tabular}{rrr|lrr|lr|lr|lr}
\toprule
\multicolumn{3}{c}{Instance} &
\multicolumn{3}{c}{A$^*$ search} &
\multicolumn{2}{c}{BS-ub} &
\multicolumn{2}{c}{BS-prob} &
\multicolumn{2}{c}{LBS} \\
 \cmidrule(lrr){1-3}\cmidrule(lrr){4-6}\cmidrule(lr){7-8} \cmidrule(lr){9-10} \cmidrule(lr){11-12} 
 $k$ & $|p_0|$  & $|\Sigma|$ & $\overline{|s_{best}|}$ & $\overline{t}[s]$ & $\overline{ub}$ & $\overline{|s_{best}|}$& $\overline{t}[s]$ & $\overline{|s_{best}|}$ & $\overline{t}[s]$ & $\overline{|s_{best}|}$ & $\overline{t}[s]$\\ \hline
\midrule
3 & 5 & 4 & \textbf{247.0} & 104.60 & 247.00 & 245.40 & 8.25 & 246.40 & 8.65 & 246.00 & 6.97 \\
3 & 5 & 20 & 97.00 & 595.77 & 131.00 & 118.20 & 78.47 & 118.40 & 85.78 & \textbf{118.6} & 22.64 \\
3 & 10 & 4 & 187.20 & 361.00 & 291.00 & \textbf{256.0} & 8.83 & \textbf{256.0} & 8.77 & \textbf{256.0} & 8.55 \\
3 & 10 & 20 & 95.20 & 595.75 & 132.80 & 117.40 & 31.03 & 117.60 & 31.05 & \textbf{118.2} & 27.60 \\
3 & 25 & 4 & 136.80 & 283.00 & 325.80 & 268.80 & 8.85 & 269.20 & 9.05 & \textbf{275.0} & 11.35 \\
3 & 25 & 20 & 95.00 & 435.00 & 130.40 & 116.80 & 26.14 & 116.80 & 24.64 & \textbf{117.2} & 28.78 \\ \hline
5 & 5 & 4 & \textbf{237.8} & 111.63 & 237.80 & 236.80 & 7.64 & 237.40 & 8.16 & 237.20 & 6.72 \\
5 & 5 & 20 & 64.20 & 595.00 & 148.80 & 113.80 & 103.53 & 115.00 & 110.09 & \textbf{115.4} & 28.76 \\
5 & 10 & 4 & 118.40 & 400.16 & 316.80 & 254.80 & 9.45 & 254.00 & 9.30 & \textbf{256.2} & 9.26 \\
5 & 10 & 20 & 68.80 & 458.21 & 144.60 & 117.20 & 39.52 & 117.60 & 40.08 & \textbf{117.8} & 35.97 \\
5 & 25 & 4 & 99.20 & 253.00 & 334.20 & 260.40 & 9.92 & 262.80 & 10.57 & \textbf{267.8} & 12.27 \\
5 & 25 & 20 & 68.80 & 420.00 & 145.80 & 115.60 & 32.14 & 118.00 & 30.76 & \textbf{118.6} & 37.06 \\ \hline
10 & 5 & 4 & \textbf{220.2} & 65.74 & 220.20 & 220.00 & 7.92 & 220.00 & 7.78 & 220.00 & 5.30 \\
10 & 5 & 20 & 45.20 & 594.00 & 158.20 & 114.00 & 121.35 & \textbf{116.4} & 134.52 & 116.20 & 36.24 \\
10 & 10 & 4 & 65.60 & 374.00 & 330.80 & 244.80 & 9.89 & 247.00 & 9.76 & \textbf{252.2} & 9.11 \\
10 & 10 & 20 & 46.00 & 334.00 & 155.60 & 113.60 & 45.38 & 115.80 & 66.76 & \textbf{117.0} & 50.64 \\
10 & 25 & 4 & 87.20 & 212.00 & 341.20 & 262.40 & 10.93 & 261.20 & 10.64 & \textbf{266.2} & 12.53 \\
10 & 25 & 20 & 44.80 & 317.00 & 157.40 & 113.00 & 41.47 & \textbf{118.4} & 43.13 & 118.00 & 50.67 \\ \hline
\bottomrule
\end{tabular}
\end{table}
\begin{table}
\caption{Results for benchmark set \textsc{Random}: $n=500, m=10$}
\label{tab:random-n-500-m-10}
\centering
\begin{tabular}{rrr|lrr|lr|lr|lr}
\toprule
\multicolumn{3}{c}{Instance} &
\multicolumn{3}{c}{A$^*$ search} &
\multicolumn{2}{c}{BS-ub} &
\multicolumn{2}{c}{BS-prob} &
\multicolumn{2}{c}{LBS} \\
 \cmidrule(lrr){1-3}\cmidrule(lrr){4-6}\cmidrule(lr){7-8} \cmidrule(lr){9-10} \cmidrule(lr){11-12} 
 $k$ & $|p_0|$  & $|\Sigma|$ & $\overline{|s_{best}|}$ & $\overline{t}[s]$ & $\overline{ub}$ & $\overline{|s_{best}|}$& $\overline{t}[s]$ & $\overline{|s_{best}|}$ & $\overline{t}[s]$ & $\overline{|s_{best}|}$ & $\overline{t}[s]$\\ \hline
\midrule  
 3 & 5 & 4 & 40.00 & 342.00 & 323.00 & 200.60 & 9.20 & 202.40 & 9.65 & \textbf{205.0} & 12.77 \\
3 & 5 & 20 & 16.80 & 382.00 & 137.80 & 52.40 & 29.49 & \textbf{53.8} & 36.37 & 53.40 & 22.87 \\
3 & 10 & 4 & 36.40 & 278.00 & 330.20 & 204.80 & 10.09 & \textbf{208.4} & 11.10 & 203.80 & 9.54 \\
3 & 10 & 20 & 16.60 & 378.00 & 138.00 & 52.80 & 23.00 & \textbf{54.0} & 25.14 & \textbf{54.0} & 24.34 \\
3 & 25 & 4 & 35.40 & 213.00 & 331.40 & 210.40 & 9.90 & 212.00 & 10.67 & \textbf{213.0} & 12.95 \\
3 & 25 & 20 & 16.40 & 362.00 & 137.20 & 52.20 & 22.95 & \textbf{53.2} & 24.60 & \textbf{53.2} & 24.42 \\ \hline
5 & 5 & 4 & 43.00 & 342.00 & 316.80 & 192.40 & 9.08 & \textbf{194.6} & 9.18 & 193.20 & 8.78 \\
5 & 5 & 20 & 15.80 & 360.00 & 137.80 & 52.40 & 32.67 & 53.20 & 41.49 & \textbf{53.4} & 24.49 \\
5 & 10 & 4 & 34.60 & 289.00 & 330.80 & 204.60 & 9.83 & 200.60 & 9.55 & \textbf{205.4} & 12.52 \\
5 & 10 & 20 & 15.60 & 344.00 & 139.00 & 52.20 & 23.89 & 53.20 & 25.11 & \textbf{53.4} & 25.75 \\
5 & 25 & 4 & 37.80 & 212.00 & 328.40 & 207.00 & 9.59 & \textbf{210.6} & 11.18 & 210.40 & 13.35 \\
5 & 25 & 20 & 16.00 & 348.00 & 139.00 & 52.60 & 23.86 & \textbf{53.8} & 25.07 & \textbf{53.8} & 25.51 \\ \hline
10 & 5 & 4 & 57.40 & 370.00 & 301.40 & 186.20 & 8.10 & \textbf{186.4} & 8.46 & 183.80 & 7.74 \\
10 & 5 & 20 & 15.80 & 310.00 & 140.40 & 52.60 & 43.93 & \textbf{53.4} & 55.23 & \textbf{53.4} & 24.38 \\
10 & 10 & 4 & 34.80 & 281.00 & 329.40 & 200.80 & 9.80 & \textbf{204.0} & 10.74 & \textbf{204.0} & 12.68 \\
10 & 10 & 20 & 14.80 & 302.00 & 141.40 & 52.80 & 26.12 & \textbf{54.0} & 25.61 & 53.80 & 27.30 \\
10 & 25 & 4 & 35.20 & 185.00 & 330.00 & 205.80 & 11.18 & 209.60 & 12.01 & \textbf{209.8} & 14.59 \\
10 & 25 & 20 & 15.40 & 304.07 & 140.00 & 52.40 & 24.75 & \textbf{53.6} & 25.77 & \textbf{53.6} & 26.32 \\ \hline 
\bottomrule
\end{tabular}
\end{table}
\begin{table}
\caption{Results for benchmark set \textsc{Random}: $n=1000, m=3$}
\label{tab:random-n-1000-m-3}
\centering
\begin{tabular}{rrr|lrr|lr|lr|lr}
\toprule
\multicolumn{3}{c}{Instance} &
\multicolumn{3}{c}{A$^*$ search} &
\multicolumn{2}{c}{BS-ub} &
\multicolumn{2}{c}{BS-prob} &
\multicolumn{2}{c}{LBS} \\
 \cmidrule(lrr){1-3}\cmidrule(lrr){4-6}\cmidrule(lr){7-8} \cmidrule(lr){9-10} \cmidrule(lr){11-12} 
 $k$ & $|p_0|$  & $|\Sigma|$ & $\overline{|s_{best}|}$ & $\overline{t}[s]$ & $\overline{ub}$ & $\overline{|s_{best}|}$& $\overline{t}[s]$ & $\overline{|s_{best}|}$ & $\overline{t}[s]$ & $\overline{|s_{best}|}$ & $\overline{t}[s]$\\ \hline
\midrule
3 & 10 & 4 & 191.20 & 368.00 & 651.20 & 502.40 & 16.74 & 505.00 & 17.69 & \textbf{507.4} & 16.55 \\
3 & 10 & 20 & 96.60 & 595.00 & 317.60 & 234.20 & 113.87 & \textbf{238.4} & 149.54 & 237.00 & 53.04 \\
3 & 20 & 4 & 135.80 & 324.00 & 680.60 & 509.60 & 18.74 & 512.60 & 18.47 & \textbf{522.2} & 17.26 \\
3 & 20 & 20 & 90.40 & 454.00 & 320.00 & 237.40 & 64.38 & 240.80 & 65.68 & \textbf{241.0} & 69.08 \\
3 & 50 & 4 & 158.80 & 253.00 & 689.40 & 535.80 & 19.91 & 539.20 & 19.76 & \textbf{550.4} & 24.44 \\
3 & 50 & 20 & 98.80 & 467.19 & 316.00 & 236.20 & 60.39 & 241.00 & 59.27 & \textbf{242.2} & 68.62 \\ \hline
5 & 10 & 4 & 123.00 & 391.00 & 673.40 & 467.20 & 16.93 & 470.40 & 17.40 & \textbf{505.6} & 17.01 \\
5 & 10 & 20 & 70.80 & 443.00 & 327.60 & 230.00 & 168.07 & \textbf{237.6} & 186.22 & 236.80 & 68.37 \\
5 & 20 & 4 & 94.40 & 314.00 & 693.00 & 501.20 & 20.64 & 503.00 & 19.50 & \textbf{513.2} & 18.88 \\
5 & 20 & 20 & 69.40 & 425.00 & 330.20 & 230.60 & 70.69 & 238.00 & 75.00 & \textbf{240.4} & 81.20 \\
5 & 50 & 4 & 117.40 & 248.00 & 690.60 & 519.40 & 20.49 & 526.00 & 19.95 & \textbf{541.8} & 24.80 \\
5 & 50 & 20 & 72.20 & 425.00 & 330.40 & 235.80 & 71.91 & 239.60 & 72.77 & \textbf{240.6} & 86.11 \\ \hline
10 & 10 & 4 & 68.80 & 400.20 & 688.00 & 460.00 & 17.34 & 464.40 & 17.65 & \textbf{485.4} & 17.49 \\
10 & 10 & 20 & 44.40 & 336.64 & 340.00 & 227.80 & 213.35 & \textbf{232.8} & 241.04 & 229.20 & 85.40 \\
10 & 20 & 4 & 72.40 & 223.00 & 696.80 & 469.80 & 19.43 & 505.20 & 20.98 & \textbf{508.8} & 19.80 \\
10 & 20 & 20 & 51.60 & 352.60 & 338.40 & 228.60 & 95.61 & 233.40 & 96.73 & \textbf{238.8} & 113.63 \\
10 & 50 & 4 & 92.00 & 200.00 & 696.20 & 470.40 & 19.66 & 517.80 & 22.29 & \textbf{537.6} & 29.33 \\
10 & 50 & 20 & 48.80 & 347.00 & 337.80 & 230.40 & 93.61 & 236.20 & 94.58 & \textbf{238.2} & 120.36 \\  \hline 
\bottomrule
\end{tabular}
\end{table}

\begin{table}
\caption{Results for benchmark set \textsc{Random}: $n=1000, m=10$}
\label{tab:random-n-1000-m-10}
\centering
\begin{tabular}{rrr|lrr|lr|lr|lr}
\toprule
\multicolumn{3}{c}{Instance} &
\multicolumn{3}{c}{A$^*$ search} &
\multicolumn{2}{c}{BS-ub} &
\multicolumn{2}{c}{BS-prob} &
\multicolumn{2}{c}{LBS} \\
 \cmidrule(lrr){1-3}\cmidrule(lrr){4-6}\cmidrule(lr){7-8} \cmidrule(lr){9-10} \cmidrule(lr){11-12} 
 $k$ & $|p_0|$  & $|\Sigma|$ & $\overline{|s_{best}|}$ & $\overline{t}[s]$ & $\overline{ub}$ & $\overline{|s_{best}|}$& $\overline{t}[s]$ & $\overline{|s_{best}|}$ & $\overline{t}[s]$ & $\overline{|s_{best}|}$ & $\overline{t}[s]$\\ \hline
\midrule
3 & 10 & 4 & 36.20 & 279.00 & 685.80 & 387.80 & 16.91 & 412.20 & 19.39 & \textbf{413.0} & 24.09 \\
3 & 10 & 20 & 16.80 & 352.00 & 320.40 & 108.60 & 54.65 & \textbf{111.0} & 69.50 & 110.00 & 59.45 \\
3 & 20 & 4 & 36.40 & 215.00 & 685.60 & 409.20 & 20.77 & 402.20 & 21.46 & \textbf{411.2} & 22.71 \\
3 & 20 & 20 & 16.60 & 395.00 & 321.80 & 108.60 & 52.49 & 111.00 & 56.20 & \textbf{111.4} & 62.42 \\
3 & 50 & 4 & 38.80 & 216.00 & 685.60 & 424.40 & 21.20 & \textbf{429.8} & 23.14 & 427.40 & 28.76 \\
3 & 50 & 20 & 16.80 & 369.00 & 320.20 & 107.80 & 52.56 & 110.40 & 57.99 & \textbf{111.0} & 60.76 \\ \hline
5 & 10 & 4 & 35.80 & 273.00 & 685.40 & 406.00 & 20.21 & 400.60 & 21.57 & \textbf{409.2} & 26.62 \\
5 & 10 & 20 & 16.00 & 344.00 & 322.80 & 108.20 & 61.20 & \textbf{111.4} & 85.40 & 110.60 & 60.84 \\
5 & 20 & 4 & 36.60 & 212.00 & 686.40 & 395.20 & 19.25 & 414.60 & 22.65 & \textbf{416.4} & 27.00 \\
5 & 20 & 20 & 16.60 & 348.00 & 321.80 & 109.80 & 54.66 & 111.40 & 58.55 & \textbf{112.0} & 64.55 \\
5 & 50 & 4 & 38.40 & 220.00 & 685.00 & 415.00 & 20.29 & 427.40 & 22.75 & \textbf{427.8} & 29.68 \\
5 & 50 & 20 & 16.00 & 338.00 & 321.40 & 107.60 & 53.71 & 110.40 & 58.35 & \textbf{110.8} & 64.80 \\ \hline
10 & 10 & 4 & 34.40 & 279.00 & 687.40 & 376.80 & 17.09 & 395.00 & 21.12 & \textbf{405.8} & 22.05 \\
10 & 10 & 20 & 15.80 & 299.00 & 325.00 & 108.20 & 65.02 & \textbf{110.8} & 77.61 & 110.60 & 62.35 \\
10 & 20 & 4 & 35.60 & 195.00 & 689.00 & 405.60 & 21.39 & \textbf{414.2} & 24.91 & 413.20 & 26.23 \\
10 & 20 & 20 & 15.20 & 291.00 & 324.40 & 108.80 & 58.61 & 111.60 & 60.03 & \textbf{111.8} & 64.85 \\
10 & 50 & 4 & 34.20 & 201.00 & 687.60 & 408.40 & 22.07 & 409.20 & 22.52 & \textbf{427.0} & 31.77 \\
10 & 50 & 20 & 15.40 & 288.00 & 324.40 & 108.20 & 56.60 & 110.40 & 58.99 & \textbf{110.8} & 61.11 \\ \hline 
\bottomrule
\end{tabular}
\end{table}

\subsection{Numerical results: benchmark set ABSTRACT}

Numerical results of the four approaches regarding dataset \textsc{Abstract} are provided in  Tables~\ref{tab:abstract-pos}--\ref{tab:abstract-neg}. These tables are organized as follows: the first two columns provide the number of input strings and the number of instances in the instance group, respectively. The following four sections present the results for A$^*$ search, BS-ub, BS-prob, and LBS, respectively. For each approach, the tables report the average solution quality and average runtime across all instances in each group. Additionally, the A$^*$ search results include the average upper bound. \\

\noindent The following conclusions can be drawn from the reported results. 
\begin{itemize}
	\item \textbf{POS instances}: LBS demonstrates superior performance compared to the other three methods for $m\leq 10$. The second-best approach is BS-prob, followed by BS-ub. A$^*$ search is the weakest performer, failing to find optimal results for any of the 149 problem instances. For instances with $m \in \{11,12\}$, BS-prob emerges as the best performer, followed by LBS, with the other two methods trailing significantly. The enhanced performance of BS-prob over LBS for larger $m$ can be attributed to the strength of the probability-based model as the number of input strings increases. It is known that the upper bound (UB) guidance, which contributes to this model, performs well when there are many similar input strings. This allows for selecting a more accurate and relevant subset of extensions $V'_{ext}$, which is crucial for constructing effective probabilistic guidance. Conversely, the neural network (NN) guidance may require more intensive training for larger instances. It could be, for example, that the utilized stopping criteria for the training process were overly strict, particularly when aiming for the best solutions in more challenging cases. Exploring various stopping criteria for the NN training process is a direction for future work. 
	
	\item \textbf{NEG instances}: in this case, LBS demonstrates superior performance only for $m \leq 7$. In contrast, for $m\geq 8$, the best-performing method is BS-prob. The strong performance of LBS in cases with smaller $m$ can be attributed to the well-trained NN, which appears to produce high-quality local solutions upon reaching the training process's stopping criterion. Conversely, as $m$ increases, BS-prob shows significant potential. The input strings, representing abstracts, are less similar and more independent compared to those in the POS group, as confirmed in the literature. This characteristic aligns well with the assumptions underlying the probabilistic guidance, leading to a better performance of BS-prob on the NEG type instances.  It is worth noting that while the structural similarity between input strings remains reasonably high, it still contributes to a favorable selection of the subset  $V'_{ext}$, similar to the behavior observed in the POS instances.
	
	\item In most cases, the RLCS solutions for instances in the POS set are larger than those for the corresponding instances in the NEG set, except for the smallest instances with $m=3$. This observation aligns with conclusions from the literature that classify  these instances to either positive (POS) or negative (NEG) group according to similarity of abstracts that each instance consists to. Furthermore, the prohibition of the 60 most frequent academic words from the final solution as a subsequence results in a final similarity score that is approximately 2\% lower than the best scores obtained when the restricted strings are omitted from the instances; see~\cite{nikolic2021solving}).   
	
\end{itemize}

\begin{table}[!t]
\caption{Results for the benchmark set \textsc{Abstract}: POS type instances}  
\label{tab:abstract-pos}
\centering
\begin{tabular}{rr|rrr|rr|lr|lr}
\hline
\multicolumn{2}{c}{\emph{Inst}} &
\multicolumn{3}{c}{A$^*$ search} &
\multicolumn{2}{c}{BS-ub}        &
\multicolumn{2}{c}{BS-prob}      &
\multicolumn{2}{c}{LBS} \\
 \cmidrule(ll){1-2}\cmidrule(lrr){3-5}\cmidrule(lr){6-7} \cmidrule(lr){8-9} \cmidrule(lr){10-11} 

$m$ & \#inst & $\overline{|s_{best}|}$ & $\overline{t}[s]$ & $\overline{ub}$ & $\overline{|s_{best}|}$ &  $\overline{t}[s]$ & $\overline{|s_{best}|}$ & $\overline{t}[s]$ & $\overline{|s_{best}|}$ & $\overline{t}[s]$ \\ \hline
\midrule
3 & 10 & 49.40 & 574.91 & 359.60 & 225.40 & 677.66 & 225.20 & 649.78 & \textbf{246.5} & 150.53 \\
4 & 10 &  42.00 & 573.60 & 353.90 & 204.20 & 592.52 & 211.60 & 604.41 & \textbf{225.5} & 124.65 \\
5 &  10 & 35.40 & 582.50 & 349.70 & 185.50 & 507.46 & 188.60 & 540.93 & \textbf{203.8} & 110.23 \\
6 &  10 & 31.20 & 582.50 & 327.50 & 165.50 & 491.03 & 171.40 & 507.56 & \textbf{178.5} & 97.64 \\
7 & 10 &  30.10 & 594.50 & 319.30 & 156.30 & 444.75 & 162.20 & 492.51 & \textbf{168.5} & 95.84 \\
8 & 10 & 27.40 & 580.28 & 304.20 & 144.30 & 419.34 & 151.50 & 467.62 & \textbf{154.1} & 95.97 \\
9 & 10 &  24.90 & 571.36 & 303.10 & 140.50 & 401.60 & 146.70 & 435.61 & \textbf{149.7} & 90.67 \\
10 & 66 & 30.12 & 592.12 & 265.36 & 126.85 & 357.50 & 136.23 & 377.25 & \textbf{137.83} & 88.49 \\
11 & 12 & 29.50 & 593.75 & 260.75 & 122.75 & 324.43 & \textbf{133.08} & 331.96 & 132.25 & 89.61 \\
12 & 1 & 29.00 & 565.00 & 257.00 & 121.00 & 297.18 & \textbf{131.0} & 294.85 & 129.00 & 80.55 \\ \hline
\bottomrule
\end{tabular}
\end{table}

\begin{table}[!t]
\caption{Results for the benchmark set \textsc{Abstract}: NEG type instances}
\label{tab:abstract-neg}
\centering
\begin{tabular}{rr|rrr|rr|lr|lr}
\hline
\multicolumn{2}{c}{\emph{Inst}}  &
\multicolumn{3}{c}{A$^*$ search} &
\multicolumn{2}{c}{BS-ub}        &
\multicolumn{2}{c}{BS-prob}      &
\multicolumn{2}{c}{LBS} \\
 \cmidrule(ll){1-2}\cmidrule(lrr){3-5}\cmidrule(lr){6-7} \cmidrule(lr){8-9} \cmidrule(lr){10-11} 

$m$ & \#inst & $\overline{|s_{best}|}$ & $\overline{t}[s]$ & $\overline{ub}$ & $\overline{|s_{best}|}$ & $\overline{t}[s]$ & $\overline{|s_{best}|}$ & $\overline{t}[s]$ & $\overline{|s_{best}|}$ & $\overline{t}[s]$ \\ \hline
\midrule
3 & 10 & 50.40 & 562.00 & 349.60 & 229.90 & 637.30 & 231.90 & 644.95 & \textbf{248.8} & 141.06 \\
4 & 10 & 42.20 & 585.65 & 343.20 & 203.50 & 554.20 & 210.10 & 635.06 & \textbf{222.8} & 123.69 \\
5 & 10 & 36.60 & 590.00 & 327.80 & 182.10 & 525.13 & 190.50 & 589.71 & \textbf{193.6} & 98.51 \\
6 & 10 & 36.70 & 528.53 & 305.20 & 165.60 & 415.16 & 172.60 & 460.74 & \textbf{173.3} & 94.92 \\
7 & 10 & 31.50 & 531.50 & 311.40 & 159.10 & 381.91 & 167.50 & 456.79 & \textbf{168.5} & 89.19 \\
8 & 10 & 34.10 & 564.00 & 286.50 & 141.30 & 345.64 & \textbf{149.5} & 433.96 & 148.00 & 85.06 \\
9 & 10 & 0.50 & 568.48 & 287.30 & 138.70 & 327.71 & \textbf{146.0} & 422.70 & 144.50 & 83.53 \\
10 & 66 & 39.44 & 550.29 & 233.94 & 126.18 & 281.39 & \textbf{134.05} & 343.09 & 132.50 & 79.44 \\
11 & 12 & 38.58 & 572.08 & 227.25 & 120.92 & 259.08 & \textbf{129.58} & 321.23 & 128.25 & 83.44 \\
12 & 1  & 38.00 & 600.00 & 221.00 & 115.00 & 245.96 & \textbf{126.0} & 296.43 & 124.00 & 72.82 \\ \hline
\bottomrule
\end{tabular}
\end{table}

\subsection{Statistical Analysis}\label{sec:statistics} 

\begin{figure}[ht]
	\centering
	\begin{subfigure}[b]{0.45\textwidth}
		\centering
		\includegraphics[width=\textwidth]{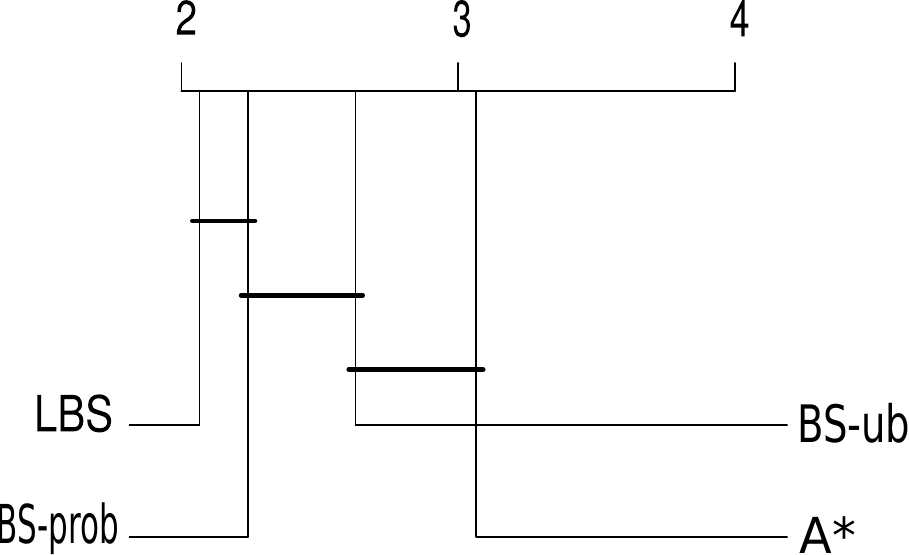}
		\caption{Instances with $n=200$}
	\end{subfigure}
	\hfill
	\begin{subfigure}[b]{0.45\textwidth}
		\centering
		\includegraphics[width=\textwidth]{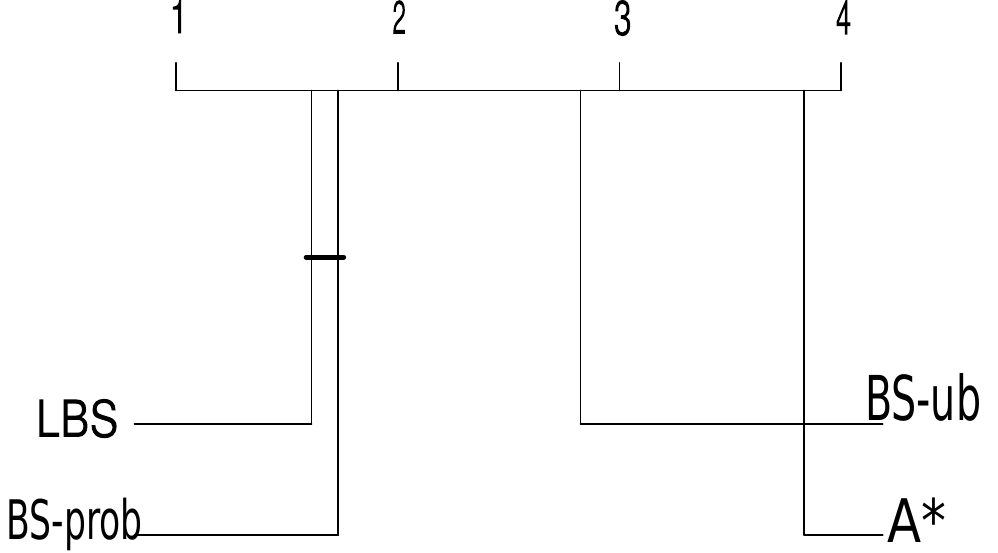}
		\caption{Instances with $n=500$}
	\end{subfigure}
	\begin{subfigure}[b]{0.45\textwidth}
		\centering
		\includegraphics[width=\textwidth]{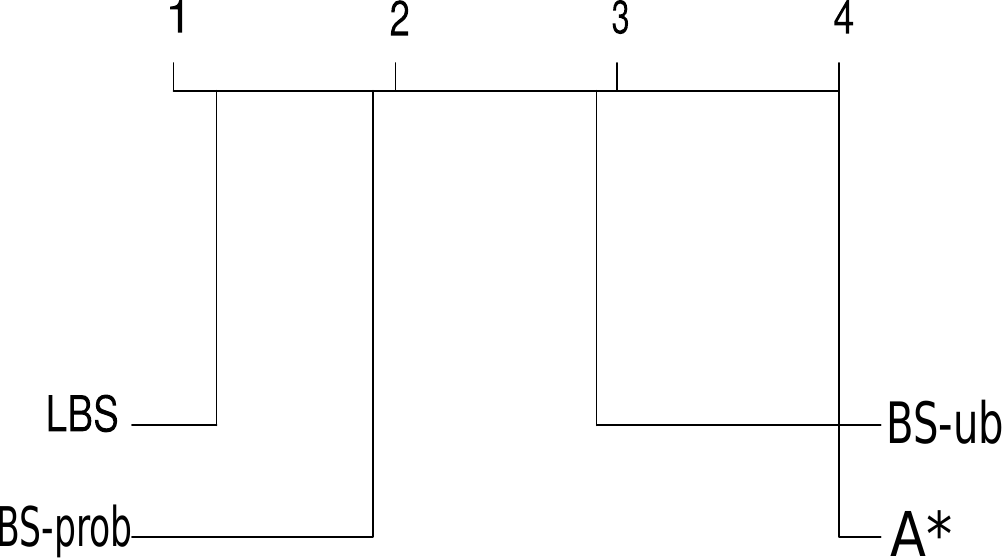}
		\caption{Instances with $n=1000$} 
	\end{subfigure}
	\begin{subfigure}[b]{0.45\textwidth}
		\centering
		\includegraphics[width=\textwidth]{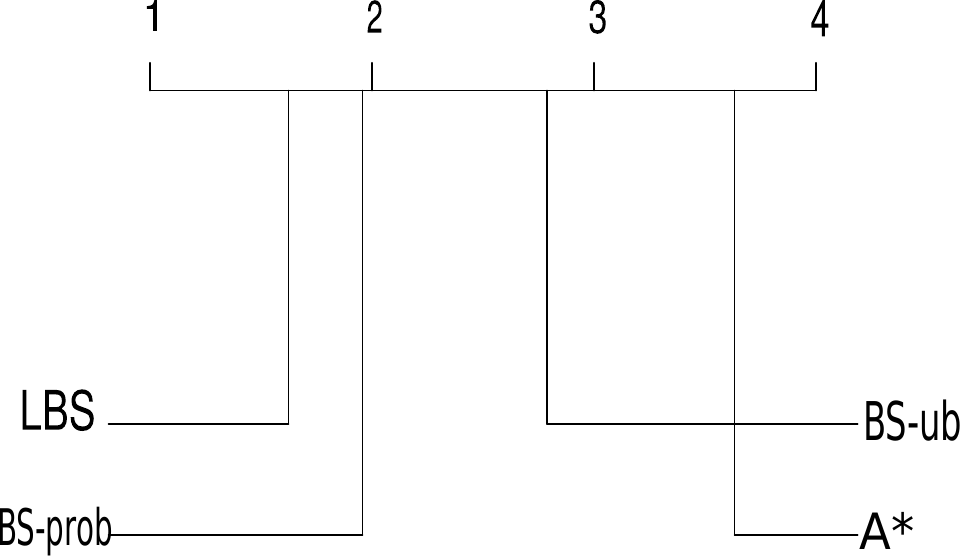}
		\caption{All instances} 
		\label{fig:cd_plots_random_combined}
	\end{subfigure}
	
	\caption{CD plots: statistical comparison for benchmark set \textsc{Random}}
	\label{fig:cd-plots-rlcs}
\end{figure}

We employ the following methodology to statistically compare the numerical results of the four algorithms. First, we categorize all our instances into two disjoint parts: random (represented by 162 groups) and real-world (represented by 20 groups). For each group the (average) solution quality is provided, obtained by executing an algorithm over the instances of that group.   
These groups from the benchmark set \textsc{Random} are further divided into three subgroups based on different values of $n$, which is the most influential characteristic; each group contains the (aggregated) results of 54 groups.  For each of the four groups, we conduct Friedman's statistical test at a significance level of 5\%, testing the null hypothesis ($H_0$) that the results of all four algorithms are statistically equivalent. If the null hypothesis is rejected, we apply the Nemenyi pairwise post-hoc test for multiple joint samples~\cite{pohlert2014pairwise}. For each pair of algorithms (6 combinations), we calculate a critical difference (CD). If the difference ($\alpha$) between the average ranks of the two algorithms is smaller than the calculated critical difference at the 5\% significance level, a horizontal bar is drawn to connect them, indicating that they perform statistically equivalently. All four algorithms are plotted along the x-axis according to their average rankings based on the results.  All four algorithms are placed at the x-axis according to the average ranking of the obtained results. The following conclusions are drawn from the obtained CD plots, plotted in Figure~\ref{fig:cd-plots-rlcs}. 

\begin{itemize}
	\item \textbf{Benchmark set \textsc{Random}}: for the instances with $n=200$, LBS achieves the highest average ranking, followed closely by BS-prob, with BS-ub and A$^*$ search trailing behind. LBS and BS-prob perform statistically equivalently, while LBS significantly outperforms both BS-ub and A$^*$ search; conversely, BS-prob is statistically equal to BS-ub. For the middle-sized instances with $n=500$, LBS again delivers the best average ranking. It performs statistically equally to BS-prob, both significantly outperforming the other two approaches. A$^*$ search is the weakest performer, as these instances are too challenging to be solved exactly. Finally, LBS is the best performer---with statistical significance--for the large instances with $n=1000$; see Figure\ref{fig:cd_plots_random_combined}. Therefore, we conclude that the LBS approach establishes a new state-of-the-art method for solving random RLCS problem instances.
	
	\item \textbf{Benchmark set \textsc{Abstract}}: the LBS approach achieves the highest average ranking, followed closely by BS-prob. However, no significant difference is observed between these two top-performing methods. In contrast, both LBS and BS-prob are significantly better than the other two approaches.
\end{itemize}

\begin{figure}[!t]
	\centering
	\includegraphics[width=200pt,height=100pt]{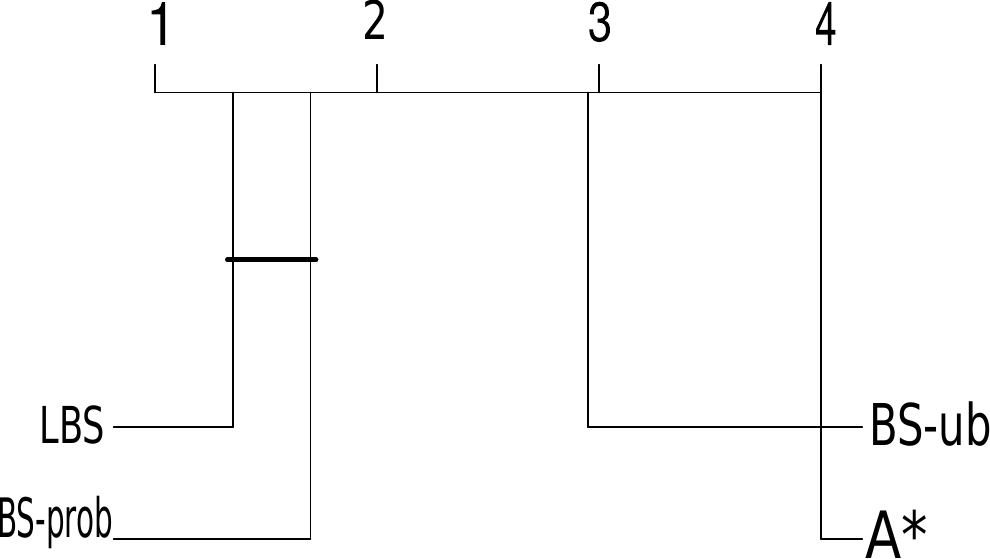}
	\caption{CD plot for benchmark set \textsc{Abstract}}    
	\label{fig:cd-plots-abstract}
\end{figure}

\section{Explainability analysis of the Designed Search Algorithms}\label{sec:ela}

We employ explainable algorithm performance prediction to provide a more comprehensive comparison of the algorithms. This analysis is focused on the \textsc{Random} benchmark set due to its well-defined instance feature set (consisting of five features). In contrast, the \textsc{Abstract} benchmark set poses challenges, such as the ambiguous meaning of the feature $n$, as the strings in $S$ are not of equal length. For this set, $n$ could refer to the minimum or average string length, complicating the analysis and necessitating the use of feature extraction algorithms, which falls outside the scope of this paper. The same issue applies to the patterns in set $P$.

The dataset is divided into training and test sets, with 25\% of the problem instances randomly selected for the test set. A single-target regression (STR) model is then trained for each algorithm separately, using the same set of instance features to predict its performance. To assess the contribution of these features, the \textit{SHapley Additive exPlanations} (SHAP)~\cite{nohara2022explanation} method is applied. SHAP calculates both global feature importance (the impact of a feature on predicting algorithm performance across the entire dataset) and local feature importance (the influence of a feature on an individual problem instance). Feature importance is computed for each model, enabling a comparative analysis of how instance features affect model predictions across different algorithms.

\subsection{Algorithm Performance Prediction}\label{subsec:performance_prediction} 
A Random Forest (RF) model~\cite{smith2013comparison} is trained to predict algorithm performance on the \textsc{Random} benchmark set. We evaluate the model's performance both with default hyperparameters and after conducting hyperparameter optimization using the grid search algorithm~\cite{alibrahim2021hyperparameter}. The results are compared against a baseline model, which simply predicts the mean performance from the training dataset. Table~\ref{tab:random-rf-cv} details the predictive capability of the models by displaying performance results, obtained with 5-fold cross-validation on the train dataset. Table~\ref{tab:random-rf-test} details the predictive capability of the models by displaying performance results on the test dataset. 

\begin{table}
\caption{Cross-validation accuracy of the RF model when predicting the performance of the algorithms on the~\textsc{Random} benchmark set.}
\label{tab:random-rf-cv}
\centering
\begin{tabular}{llrr}
\toprule
{} & algorithm &         MAE &        R$^2$ \\
model    &           &             &           \\
\midrule
baseline &        A* &   26.500068 & -0.020399 \\
 &     BS-ub &  125.790807 & -0.009682 \\
 &   BS-prob &  127.061167 & -0.009730 \\
 &       LBS &  129.727857 & -0.008810 \\ \hline \hline
RF\_default &        A* &  7.565803 &  0.654524 \\
 &     BS-ub &  4.673359 &  0.997649 \\
 &   BS-prob &  4.926889 &  0.997348 \\
 &       LBS &  {3.619643} &  0.998527 \\
\bottomrule
\end{tabular}
\end{table}

The predictive performance of the models is evaluated using two metrics: Mean Absolute Error (MAE) and the coefficient of determination (R$^2$ score). Lower MAE values and higher R$^2$ scores indicate better performance. The results demonstrate that all RF models perform well in predicting algorithm performance, significantly outperforming the baseline model. Notably, the RF model with default hyperparameters achieves the best results across both metrics, highlighting its superior predictive accuracy.

\begin{table}
\caption{Test accuracy of the RF model when predicting the performance of the algorithms on the~\textsc{Random} benchmark set.}
\label{tab:random-rf-test}
\centering
\begin{tabular}{llrr}
\toprule
{} & algorithm &         MAE &        R$^2$ \\
model    &           &             &           \\
\midrule
baseline &        A$^*$ &   35.189921 & -0.009718 \\
 &     BS-ub &  101.473614 & -0.083254 \\
 &   BS-prob &  102.814191 & -0.080225 \\
 &       LBS &  105.217738 & -0.076159 \\
 \midrule \hline
RF\_default &        A$^*$ &  12.874537 &  0.646386 \\
 &     BS-ub &   5.062878 &  0.994857 \\
 &   BS-prob &   5.349073 &  0.992730 \\
 &       LBS &   4.271854 &  0.996434 \\
\bottomrule
\end{tabular}
\end{table}

\subsection{Model Explainability Analysis}

\begin{figure}[htbp]
	\centering
	\begin{subfigure}[b]{0.45\textwidth}
		\centering        \includegraphics[width=\textwidth]{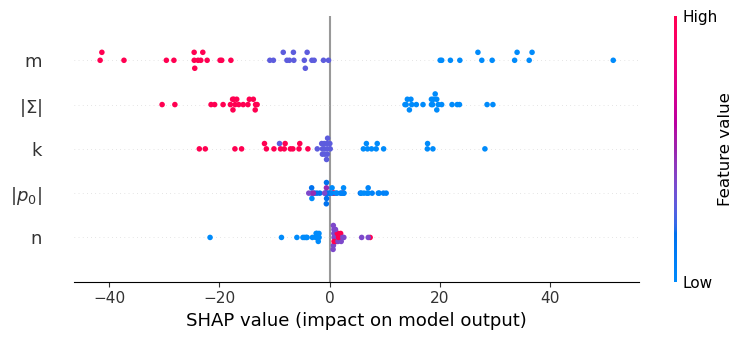}
		\caption{SHAP summary plot for A$^*$ search}
	\end{subfigure}
	\hfill
	\begin{subfigure}[b]{0.45\textwidth}
		\centering
		\includegraphics[width=\textwidth]{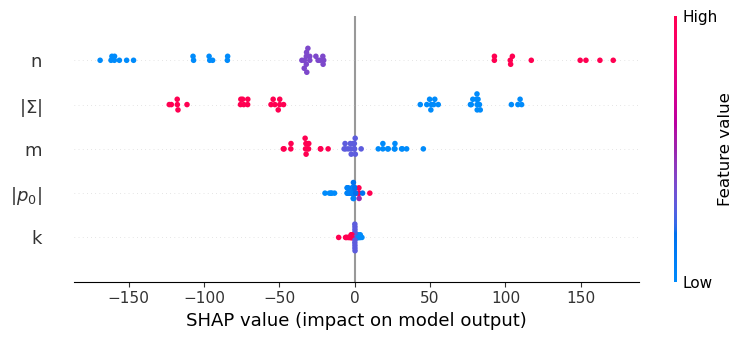}
		\caption{SHAP summary plot for BS-ub}    \end{subfigure}
	\hfill
	\begin{subfigure}[b]{0.45\textwidth}
		\centering
		\includegraphics[width=\textwidth]{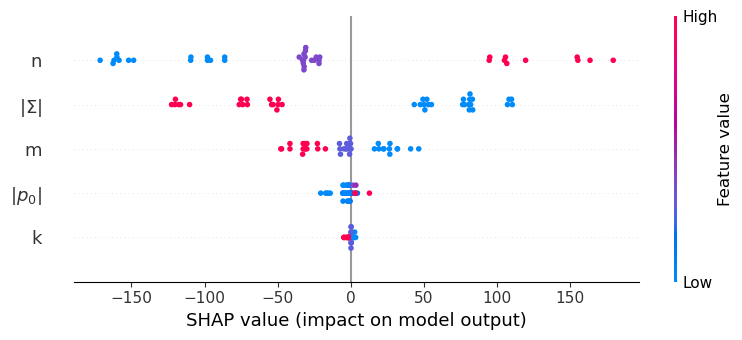}
		\caption{SHAP summary plot for BS-prob}    \end{subfigure}
	\hfill
	\begin{subfigure}[b]{0.45\textwidth}
		\centering
		\includegraphics[width=\textwidth]{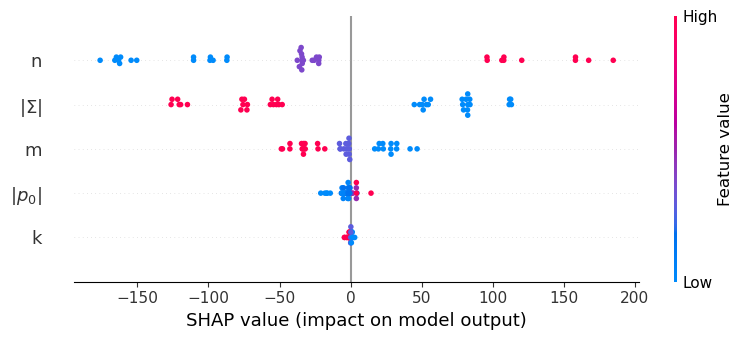}
		\caption{SHAP summary plot for LBS}    \end{subfigure}
	\caption{SHAP summary plots: Shows the global feature importance and the direction of influence of different feature values (i.e. low in blue to high in red)  on the~\textsc{Random} benchmark set. }
	\label{fig:summary-plots-rlcs}
\end{figure}

The SHAP global feature importance for A$^*$, BS-ub, BS-prob, and LBS is illustrated in Figure~\ref{fig:summary-plots-rlcs}, with each subfigure corresponding to one algorithm. The y-axis lists the instance features in descending order of their global importance. Each point on the plot represents a problem instance, and the horizontal  position of the point reflects the importance of the feature--points farther to the right or left indicate greater importance. The vertical gray line at 0 represents the model's average prediction and serves as a reference point, meaning features near this line have no significant impact on the model's output. A point positioned to the right (positive value) indicates that the feature increases the model's prediction, suggesting better algorithm performance. Conversely, a point to the left (negative value) indicates that the feature reduces the prediction, implying worse algorithm performance. The color gradient of the points, ranging from dark blue (low feature values) to yellow (high feature values), further explains how feature values contribute to the effect. Features consistently positioned far to the left or right (i.e., high importance) are generally impactful across all instances. In contrast, points clustered near 0 indicate minimal overall importance. If the points for a feature are widely dispersed, this suggests that its influence varies significantly across different problem instances.


From the plots, one can see that for the A$^*$ algorithm, the number of input strings ($m$) has the highest impact on the model's output on average. Low values of $m$ (in blue) tend to contribute to longer predicted subsequences. Contrary, high values of $m$ (in red) contribute to  predicting shorter subsequences. This is as expected as with a small number of input strings, the search space gets smaller, and it is generally easier to find a longer common subsequence that avoids restricted patterns. With more input strings, the space of common subsequences gets smaller thus less candidates for the restricted common subequences is expected.   

The size of the alphabet ($|\Sigma|$) also has a considerable impact as the second most important feature on average. Higher alphabet sizes (in red) tend to impact the model's output, suggesting shorter common subsequences. Lower alphabet sizes (in blue) lead to longer predicted subsequences. It is commonly acknowledged that larger alphabet sizes tend to increase problem complexity by introducing more distinct characters and potential subsequences, making it harder to find valid common subsequences while avoiding restricted patterns. 
In contrast, a smaller alphabet reduces conflicting patterns and search space, allowing the model more flexibility to find longer valid subsequences. 

The number of restricted strings ($k$) impacts the complexity of the RLCS problem, though its effect is not always as pronounced as that of $n$ or $|\Sigma|$. As $k$ increases, the model must avoid a greater number of restricted patterns, which can shorten the predicted common subsequence. However, the difficulty posed by restricted strings can vary. Some patterns may occur infrequently in the input strings, or the solution space might naturally circumvent them. In such cases, even with a larger $k$, the model may still find longer subsequences, reducing the overall influence of $k$. Other features, such as the length of restricted strings ($|p_0|$) and the length of the input strings ($n$), show minimal to no impact on the model's predictions, as evidenced by the dense clustering of points near zero in the SHAP plots. 

For the remaining algorithms—BS-ub, BS-prob, and LBS—the feature importance plots are nearly identical, with the length of the input strings ($n$) emerging as the most influential factor. The alphabet size ($|\Sigma|$) also plays a significant role in shaping the model's predictions, similar to findings reported for the basic LCS problem~\cite{djukanovic2020finding}. In comparison, the number of input strings ($m$) has a more moderate impact on the model's output, though it remains less influential than $n$ and $|\Sigma|$. As with A$^*$ search, the number of restricted strings ($k$) and the length of these restricted strings ($|p_0|$) have a relatively minor effect on the BS-based algorithms. This reduced influence of $k$ and $p_0$ is likely because there are many restricted common subsequences in the solution space with lengths similar to that of the basic LCS (without considering restricted strings). As a result, the BS-based algorithms are effective in locating these sequences. This also explains why the basic upper bound for the LCS problem performs well as a guide for navigating the RLCS problem space.

We can note a significant difference between A$^*$ and the BS algorithms in terms of feature importance. As an exact algorithm, A$^*$ search is inherently constrained by the NP-hard nature of the problem, which leads to scalability issues and poor anytime performance as the problem size increases~\cite{djukanovic2020finding}. Specifically, once a critical instance size is exceeded (in this case, $n \geq 500$), the (sub-optimal) performance of A$^*$ is expected, no longer primarily influenced by $n$; instead, other features become more relevant as the algorithm tends to stagnate at top (i.e. upper) levels of the search tree, reverting to breadth-first-search iterations.

In contrast, the BS algorithms, being metaheuristic approaches controlled by the beam width parameter, are designed for robustness and scalability, ensuring better anytime performance. At each level, BS improves partial solutions incrementally, guiding the search towards deeper levels and higher-quality complete solutions. Here, the quality of solutions, represented by the depth of the search tree reached, is directly dependent on $n$—larger values of $n$ are associated with higher-quality solutions.\\

Next, we choose several specific problem instances corresponding to the different cases pointed out in Section~\ref{subsection:results_random} from our dataset to illustrate and compare the \textit{local} feature importance across the algorithms, and in this way exposing similarities and differences in algorithm behavior based on instance features. 

\begin{figure}[htbp]
	\centering
	\begin{subfigure}[b]{0.45\textwidth}
		\centering
		\includegraphics[height=120pt]{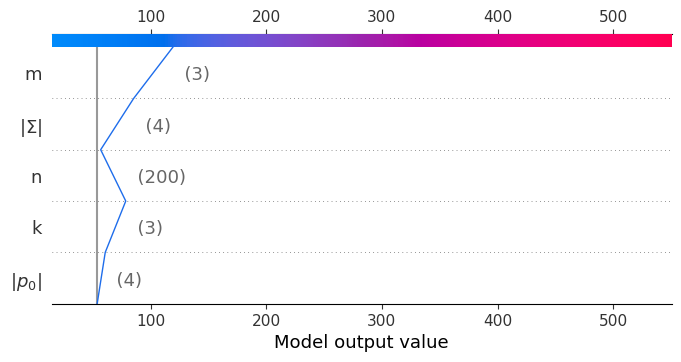}
		\caption{Decision plot for A$^*$.}
	\end{subfigure}
	\hfill
	\begin{subfigure}[b]{0.45\textwidth}
		\centering
		\includegraphics[height=120pt]{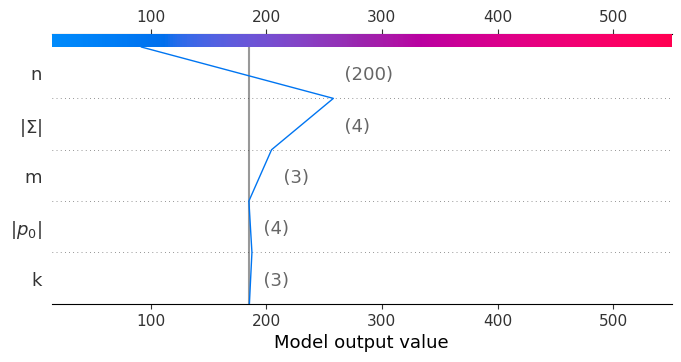}
		\caption{Decision plot for LBS.}    \end{subfigure}
	\caption{SHAP decision plots on the group $m=3, n=200, k=3, 
		|p_0|=4, |\Sigma|=4$ of the \textsc{Random} benchmark set (averaged over 5 instances).}
	\label{fig:decision-plots-rlcs-small}
\end{figure}

In the case of small-sized instances  with $n=200$, the SHAP local feature importance for A$^*$ and LBS is presented in Figure~\ref{fig:decision-plots-rlcs-small} for the problem instance group $m=3, n=200, k=3, p_0=4$, and $|\Sigma|=4$, averaged over its  5 instances. To maintain clarity, we focus on A$^*$ and LBS plots, as the visualizations for BS-ub and BS-prob are nearly identical to those of LBS. The y-axis lists the features in descending order of importance, with the most influential feature at the top, while the x-axis reflects the model's predicted values.  Each line in the plot corresponds to a unique problem instance (five in total for this group) and represents the cumulative sum of the local feature importance, tracing a path from the model's base value to its final prediction. The base value, shown as a vertical gray line, denotes the average algorithm performance across the training dataset. The final intersection of each line with the x-axis indicates the model's predicted value for that instance, with line colors transitioning from dark blue (low values) to red (high values) to represent predicted performance levels. The direction of each line's deviation at a given feature point indicates that feature's influence on the model's prediction. A shift to the right suggests that the feature increases the predicted performance, while a shift to the left implies a reduction in performance. Numbers in parentheses next to the line segments denote the specific feature values for each instance.

For small-sized instances, all the algorithms were able to effectively find the optimal solution in majority of the cases as reported in Table~\ref{tab:random-n-200-m-3}. We can see that the two models corresponding to A$^*$ and LBS come to similar predicted solution (around 100), however due to different reasons. 
The features $m$ and $|\Sigma|$, $n$ and $k$ have a very balanced influence on the prediction for A$^*$ search. The prediction for LBS depends heavily on $n$, followed by $|\Sigma|$ and $m$, while the factors related to the restricted strings have no a significant impact. 

\begin{figure}[htbp]
	\centering
	\begin{subfigure}[b]{0.45\textwidth}
		\centering
		\includegraphics[height=120pt]{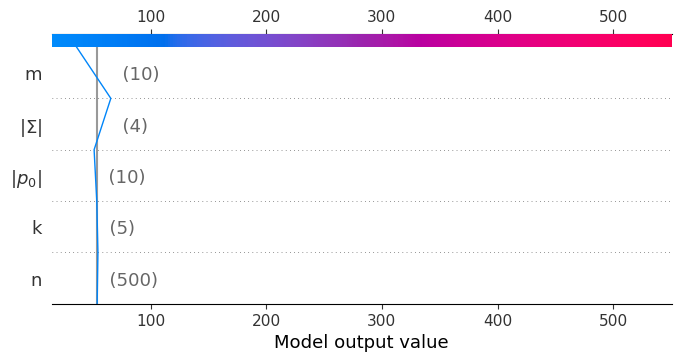}
		\caption{Decision plot for A$^*$ search.}
	\end{subfigure}
	\begin{subfigure}[b]{0.45\textwidth}
		\centering
		\includegraphics[height=120pt]{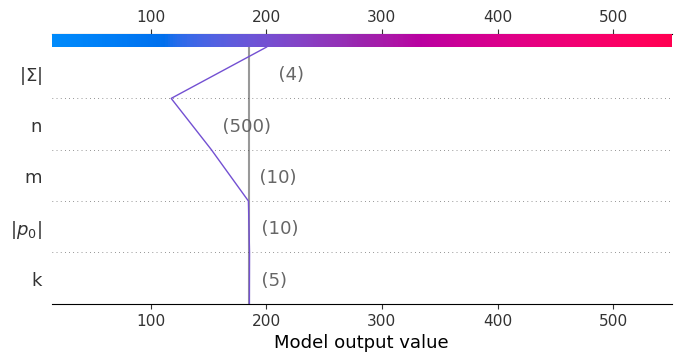}
		\caption{Decision plot for LBS.}    \end{subfigure}
	\caption{SHAP decision plots on the group $n=500, m=10, k=5, |p_0|=10, |\Sigma|=4$ of the \textsc{Random} benchmark set (averaged over 5 instances).}
	\label{fig:decision-plots-rlcs-middle}
\end{figure}

Figure~\ref{fig:decision-plots-rlcs-middle} presents a case of middle-sized problem instances (belonging to a group) with $n=500$. In this scenario, the A$^*$ search encounters significant difficulties, failing to solve most of the problem instances, see Tables~\ref{tab:random-n-500-m-3}--\ref{tab:random-n-500-m-10}. In contrast, the LBS algorithm consistently performs well across all instances. The decision plot shows that A$^*$ search performance is primarily influenced by the number of input strings ($m$) and the alphabet size ($|\Sigma|=4$), while the length of the input strings ($n$) has minimal impact on its predictions. This is likely due to the A$^*$ algorithm's suboptimal behavior when handling instances with larger $n$. As previously discussed, in such cases, A$^*$ often functions similarly to breadth-first search, rarely yielding high-quality solutions as it tends to get stuck in the upper levels of the search state graph.  In contrast, LBS's performance is largely driven by the alphabet size ($|\Sigma|$), the number of input strings ($m$), and crucially, the length of the input strings ($n$). This indicates that LBS can effectively handle the complexities of larger problem instances, demonstrating significant scalability. By factoring in both the length of the input strings and the alphabet size, LBS is able to adapt its solution strategies, resulting in successful outcomes even in more challenging scenarios.

The reason why $|\Sigma|$ has a significant impact on the performance of all algorithms is argued by the fact that  this value correlates to the branching factor of the designed search algorithms, i.e. the number of generated children of a node; the larger the cardinality of $\Sigma$, the larger the branching factor is expected for each node. Consequently, a larger number of subproblems are expected to be generated from a parent node when $|\Sigma|$ is large. However, the dimensions of these subproblems are mostly significantly reduced compared to the parent node's subproblem dimension. A smaller dimension difference is expected in the case the branching factor is lower. Hence, the larger $|\Sigma|$ contributes shorter resulting common subsequences and it holds for any of the proposed search algorithms.

\begin{figure}[htbp]
	\centering
	\begin{subfigure}[b]{0.45\textwidth}
		\centering
		\includegraphics[height=100pt]{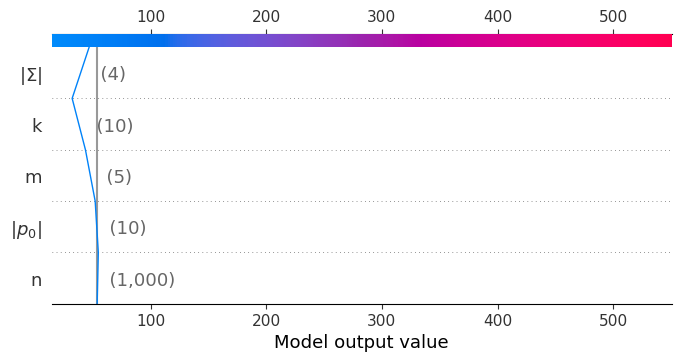}
		\caption{Decision plot for A$^*$ search.}
	\end{subfigure}
	\begin{subfigure}[b]{0.45\textwidth}
		\centering
		\includegraphics[height=100pt]{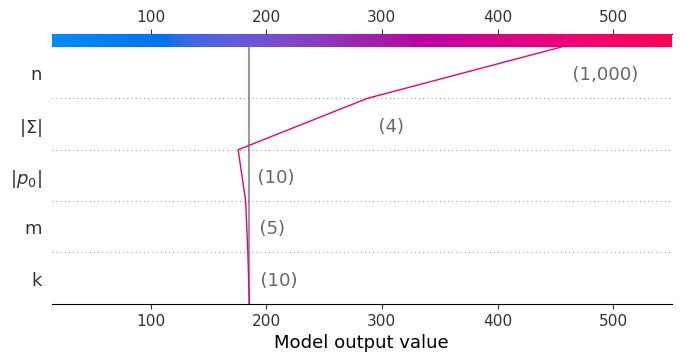}
		\caption{Decision plot for LBS.}    \end{subfigure}
	\caption{SHAP decision plots on the group $n=1000, m=5, k=10, |p_0|=10, |\Sigma|=4$ of the \textsc{Random} benchmark set (averaged over 5 instances).}
	\label{fig:decision-plots-rlcs-49}
\end{figure}

The local feature importance for the case of large-sized problem instances of an instance group with $n=1000$, for the A$^*$ search and LBS algorithms is presented in Figure~\ref{fig:decision-plots-rlcs-49}. Here, the A$^*$  algorithm fails by large amount compared to the BS algorithms, as presented in Tables~\ref{tab:random-n-1000-m-3}--\ref{tab:random-n-1000-m-10}. In contrast, the LBS algorithm proves its capability. The decision plot reveals that the performance of the A$^*$ search for the largest problem instances is, influenced by the alphabet size $|\Sigma|=4$ and the number of input strings $m$. Also, the number of restricted strings shows an influence on the performance, contributing to a shorter solution.  
Although the effect of the number of restricted strings is as expected, the A$^*$ search  algorithm's inability to balance the impact of various problem features contributes to its sub-optimal performance. The decision plot for LBS shows that its performance is influenced by a combination of features, including the length of the input strings ($n$) and the alphabet size ($|\Sigma|$) all have a significant contributions to larger solution length. This means that the LBS algorithm is able to account for all the factors influencing the problem's complexity.

\section{Conclusions and future work}\label{sec:future_work}

This paper addresses the restricted longest common subsequence problem, a generalization of the well-known longest common subsequence problem, with applications in computational biology and pattern recognition. This problem offers a nuanced measure of structural similarity between molecular structures. Firstly, a novel probability-based heuristic is introduced to guide beam search toward regions in the search space containing high-quality solutions. Secondly, a complementary heuristic is developed by training a neural network that leverages carefully selected local (subproblem-specific) and global (problem-wide) numerical features. The trained neural network evaluates the potential of each extension of a partial solution, forming the basis for a learning-based beam search framework aimed at solving the problem more effectively. To robustly assess the efficacy of these approaches, two large and diverse datasets were generated---one based on random generation and another derived from real-world scenarios---offering a broader testing ground compared to the small-scale instances available in the existing literature. The real-world dataset consists of several scientific abstracts. At the same time, the restricted strings correspond to the 60 most frequent academic words in scientific literature, allowing the investigation of their impact on the overall similarity score between abstracts. 

Our comprehensive experimental analysis revealed that the learning beam search consistently outperforms the other search approaches, particularly on large-scale random problem instances. While it also performed better than the beam search guided by probability-based heuristics on real-world problem instances, the difference between these two methods is not statistically significant. Additionally, we conducted an explainability algorithm analysis using Shapley Additive Explanations (SHAP) to support the findings from the performed statistical analysis and gain deeper insights into how individual instance features impact algorithm performance. The analysis showed that the beam search algorithms are influenced by multiple problem features, especially $m$, $n$, and $|\Sigma|$, highlighting their adaptability to varying problem sizes and complexities. In contrast, the A$^*$ algorithm's performance is predominantly influenced by the number of input strings ($m$), displaying less sensitivity to other features, which may account for its weaker performance in more complex scenarios.

Future work could explore a more general variant of the addressed problem by incorporating a set of constrained patterns into the (restricted) common subsequence, as suggested by Farhana et al.~\cite{farhana2015constrained}. This would involve combining two distinct variants of the longest common subsequence problem into a single, more complex problem, which poses additional practical and theoretical challenges.  Another potential direction is to experiment with different algorithms, beyond genetic algorithms, for offline tuning of the neural network. This could potentially enhance the performance of the current learning beam search framework. For instance, the Ant colony optimization~\cite{mavrovouniotis2015training}, a robust metaheuristic, may offer promising results and is worth investigating.

\vspace{2.0cm}
\noindent \textbf{Acknowledgments.} The research of M.~Djukanovi\'c is partially supported by the Ministry for Scientific and Technological Development and Higher Education of the Republic of Srpska, B\&H in the course of the bilateral research project between B\&H and Slovenia entitled  ``Theoretical and computational aspects of some graph problems with the application to graph network information spreading'' under project no. 1259084 and the project ``Development of models and algorithms of artificial intelligence for solving difficult combinatorial optimization problems''  under project no. 1259086. 
A.~Kartelj was supported by grant 451-03-47/2023-01/200104 funded by the Ministry of Science Technological Development and Innovations of the Republic of Serbia. J.~Reixach and C.~Blum are supported by grants TED2021-129319B-I00 and PID2022-136787NB-I00 funded by MCIN/AEI/10.13039/501100011033. The authors would like to thank the Compute Cluster Unit of the Institute of Logic and Computation at the Vienna University of Technology for providing computing resources for this research project.






  \bibliographystyle{splncs04}
 \bibliography{eswa_learning_beam_search}
 
 
 \appendix
 
 \section{Complete numerical results for the benchmark set RANDOM}\label{sec:appendix-a}
 \vspace{2cm}
  
 \begin{table}
\caption{Results for benchmark set \textsc{Random}: $n=200, m=5$}
\label{tab:random-n-200-m-5}
\centering
\begin{tabular}{rrr|lrr|lr|lr|lr}
\toprule
\multicolumn{3}{c}{Instance} &
\multicolumn{3}{c}{A$^*$ search} &
\multicolumn{2}{c}{BS-ub} &
\multicolumn{2}{c}{BS-prob} &
\multicolumn{2}{c}{LBS} \\
 \cmidrule(lrr){1-3}\cmidrule(lrr){4-6}\cmidrule(lr){7-8} \cmidrule(lr){9-10} \cmidrule(lr){11-12} 
 $k$ & $|p_0|$  & $|\Sigma|$ & $\overline{|s_{best}|}$ & $\overline{t}[s]$ & $\overline{ub}$ & $\overline{|s_{best}|}$ & $\overline{t}[s]$ & $\overline{|s_{best}|}$ & $\overline{t}[s]$ & $\overline{|s_{best}|}$ & $\overline{t}[s]$\\ \hline
\midrule
3 & 2 & 4 & \textbf{74.2} & 28.08 & 74.20 & \textbf{74.2} & 1.29 & \textbf{74.2} & 1.37 & \textbf{74.2} & 1.02 \\
3 & 2 & 20 & \textbf{29.4} & 94.71 & 29.40 & 29.20 & 18.73 & \textbf{29.4} & 19.37 & \textbf{29.4} & 4.27 \\
3 & 4 & 4 & \textbf{89.2} & 220.55 & 89.20 & 88.80 & 3.09 & 88.80 & 3.15 & \textbf{89.2} & 3.01 \\
3 & 4 & 20 & \textbf{30.0} & 120.18 & 30.00 & \textbf{30.0} & 7.40 & \textbf{30.0} & 7.92 & \textbf{30.0} & 5.46 \\
3 & 10 & 4 & 66.40 & 329.00 & 111.20 & 93.80 & 3.42 & 94.20 & 3.59 & \textbf{94.6} & 4.26 \\
3 & 10 & 20 & \textbf{29.4} & 52.66 & 29.40 & \textbf{29.4} & 5.61 & \textbf{29.4} & 5.17 & \textbf{29.4} & 5.53 \\ \hline
5 & 2 & 4 & \textbf{55.0} & 5.11 & 55.00 & \textbf{55.0} & 0.62 & \textbf{55.0} & 0.63 & \textbf{55.0} & 0.44 \\
5 & 2 & 20 & \textbf{27.6} & 225.15 & 27.60 & \textbf{27.6} & 23.47 & \textbf{27.6} & 23.75 & \textbf{27.6} & 4.77 \\
5 & 4 & 4 & \textbf{89.8} & 189.48 & 89.80 & 89.40 & 3.34 & \textbf{89.8} & 3.33 & 89.40 & 2.77 \\
5 & 4 & 20 & \textbf{29.6} & 280.09 & 29.60 & \textbf{29.6} & 9.67 & \textbf{29.6} & 10.41 & \textbf{29.6} & 6.39 \\
5 & 10 & 4 & 58.00 & 340.00 & 115.00 & 93.00 & 3.63 & \textbf{93.2} & 3.84 & \textbf{93.2} & 3.92 \\
5 & 10 & 20 & 29.60 & 177.47 & 30.80 & 29.80 & 6.22 & \textbf{30.2} & 6.21 & \textbf{30.2} & 6.50 \\ \hline
10 & 2 & 4 & \textbf{38.0} & 0.21 & 38.00 & \textbf{38.0} & 0.00 & \textbf{38.0} & 0.00 & \textbf{38.0} & 0.00 \\
10 & 2 & 20 & 27.80 & 561.17 & 32.40 & \textbf{29.2} & 28.17 & \textbf{29.2} & 27.78 & \textbf{29.2} & 5.82 \\
10 & 4 & 4 & \textbf{75.0} & 50.91 & 75.00 & 74.40 & 2.04 & 74.40 & 2.11 & 74.40 & 1.69 \\
10 & 4 & 20 & 25.60 & 595.00 & 35.80 & \textbf{29.8} & 12.63 & \textbf{29.8} & 13.46 & \textbf{29.8} & 8.63 \\
10 & 10 & 4 & 46.00 & 353.00 & 117.20 & 89.60 & 3.82 & 89.60 & 3.84 & \textbf{90.2} & 3.69 \\
10 & 10 & 20 & 26.20 & 389.00 & 36.00 & \textbf{29.6} & 8.76 & \textbf{29.6} & 8.63 & \textbf{29.6} & 9.12 \\ \hline
\bottomrule
\end{tabular}
\end{table}

 \begin{table}
\caption{Results for benchmark set \textsc{Random}: $n=500, m=5$}
\label{tab:random-n-500-m-5}
\centering
\begin{tabular}{rrr|lrr|lr|lr|lr}
\toprule
\multicolumn{3}{c}{Instance} &
\multicolumn{3}{c}{A$^*$ search} &
\multicolumn{2}{c}{BS-ub} &
\multicolumn{2}{c}{BS-prob} &
\multicolumn{2}{c}{LBS} \\
 \cmidrule(lrr){1-3}\cmidrule(lrr){4-6}\cmidrule(lr){7-8} \cmidrule(lr){9-10} \cmidrule(lr){11-12} 
 $k$ & $|p_0|$  & $|\Sigma|$ & $\overline{|s_{best}|}$ & $\overline{t}[s]$ & $\overline{ub}$ & $\overline{|s_{best}|}$ & $\overline{t}[s]$ &$\overline{|s_{best}|}$ & $\overline{t}[s]$ & $\overline{|s_{best}|}$ & $\overline{t}[s]$\\ \hline
\midrule
3 & 5 & 4 & 83.60 & 453.00 & 307.40 & 223.60 & 8.90 & \textbf{225.6} & 9.40 & 225.40 & 9.40 \\
3 & 5 & 20 & 36.00 & 595.00 & 134.60 & 79.60 & 41.85 & \textbf{80.0} & 52.52 & 79.60 & 22.50 \\
3 & 10 & 4 & 68.40 & 343.00 & 321.80 & \textbf{230.0} & 9.44 & \textbf{230.0} & 9.72 & 229.00 & 9.20 \\
3 & 10 & 20 & 36.60 & 473.00 & 131.60 & 78.00 & 23.44 & \textbf{79.0} & 22.79 & 78.80 & 25.19 \\
3 & 25 & 4 & 69.60 & 242.00 & 329.60 & 237.20 & 9.32 & 233.60 & 9.29 & \textbf{243.2} & 12.59 \\
3 & 25 & 20 & 39.00 & 501.00 & 130.00 & 79.40 & 22.57 & 79.80 & 22.61 & \textbf{80.4} & 26.35 \\ \hline
5 & 5 & 4 & 89.40 & 419.00 & 295.40 & 213.00 & 7.85 & \textbf{213.8} & 8.36 & 212.80 & 8.00 \\
5 & 5 & 20 & 32.20 & 595.00 & 137.80 & 78.80 & 61.88 & \textbf{79.0} & 68.42 & 78.80 & 25.43 \\
5 & 10 & 4 & 54.20 & 347.00 & 329.00 & 225.40 & 9.32 & \textbf{228.8} & 10.11 & 228.40 & 9.83 \\
5 & 10 & 20 & 32.80 & 441.00 & 137.00 & 78.60 & 25.29 & \textbf{80.0} & 25.79 & \textbf{80.0} & 30.39 \\
5 & 25 & 4 & 57.80 & 249.00 & 331.60 & 233.60 & 9.81 & 235.80 & 9.96 & \textbf{239.2} & 12.86 \\
5 & 25 & 20 & 33.60 & 444.00 & 137.00 & 79.00 & 25.35 & 79.00 & 25.42 & \textbf{79.4} & 29.36 \\ \hline
10 & 5 & 4 & 114.80 & 408.00 & 269.80 & 202.20 & 8.07 & \textbf{213.0} & 8.84 & 212.80 & 8.01 \\
10 & 5 & 20 & 27.40 & 589.00 & 143.40 & 77.00 & 78.45 & \textbf{78.2} & 89.87 & \textbf{78.2} & 30.74 \\
10 & 10 & 4 & 46.40 & 346.00 & 333.00 & 228.60 & 10.09 & \textbf{230.6} & 10.52 & \textbf{230.6} & 9.23 \\
10 & 10 & 20 & 24.80 & 345.00 & 145.60 & 77.80 & 32.18 & 79.20 & 31.53 & \textbf{79.4} & 37.59 \\
10 & 25 & 4 & 51.60 & 198.00 & 333.80 & 223.40 & 9.99 & 225.00 & 10.21 & \textbf{237.6} & 13.60 \\
10 & 25 & 20 & 26.00 & 359.00 & 145.20 & 78.40 & 32.12 & \textbf{79.2} & 30.26 & \textbf{79.2} & 36.05 \\ \hline 
\bottomrule
\end{tabular}
\end{table}
 \begin{table}
\caption{Results for benchmark set \textsc{Random}: $n=1000, m=5$}
\label{tab:random-n-1000-m-5}
\centering
\begin{tabular}{rrr|lrr|lr|lr|lr}
\toprule
\multicolumn{3}{c}{Instance} &
\multicolumn{3}{c}{A$^*$ search} &
\multicolumn{2}{c}{BS-ub} &
\multicolumn{2}{c}{BS-prob} &
\multicolumn{2}{c}{LBS} \\
 \cmidrule(lrr){1-3}\cmidrule(lrr){4-6}\cmidrule(lr){7-8} \cmidrule(lr){9-10} \cmidrule(lr){11-12} 
 $k$ & $|p_0|$  & $|\Sigma|$ & $\overline{|s_{best}|}$ & $\overline{t}[s]$ & $\overline{ub}$ & $\overline{|s_{best}|}$ & $\overline{t}[s]$ & $\overline{|s_{best}|}$ & $\overline{t}[s]$ & $\overline{|s_{best}|}$ & $\overline{t}[s]$\\ \hline
\midrule
3 & 10 & 4 & 65.40 & 348.00 & 682.00 & 453.60 & 18.96 & \textbf{456.8} & 20.40 & 455.40 & 18.64 \\
3 & 10 & 20 & 37.80 & 484.00 & 315.40 & 159.40 & 71.02 & \textbf{162.4} & 83.59 & \textbf{162.4} & 54.00 \\
3 & 20 & 4 & 69.20 & 270.00 & 686.20 & 459.60 & 18.07 & 462.40 & 20.08 & \textbf{463.2} & 20.80 \\
3 & 20 & 20 & 36.40 & 485.00 & 317.60 & 161.00 & 52.11 & 162.80 & 52.39 & \textbf{163.0} & 60.69 \\
3 & 50 & 4 & 69.80 & 258.00 & 685.60 & 471.40 & 19.05 & 477.20 & 21.09 & \textbf{488.0} & 24.83 \\
3 & 50 & 20 & 36.60 & 481.00 & 314.60 & 159.60 & 52.35 & 160.80 & 53.10 & \textbf{162.2} & 63.47 \\ \hline
5 & 10 & 4 & 52.80 & 349.00 & 686.80 & 448.80 & 19.82 & 440.60 & 19.25 & \textbf{451.8} & 19.85 \\
5 & 10 & 20 & 31.60 & 450.00 & 321.40 & 158.80 & 94.43 & \textbf{161.8} & 115.87 & 161.60 & 63.09 \\
5 & 20 & 4 & 59.80 & 231.00 & 687.80 & 453.40 & 18.31 & 459.00 & 19.82 & \textbf{463.8} & 23.11 \\
5 & 20 & 20 & 31.20 & 451.00 & 322.00 & 158.20 & 57.14 & 162.40 & 57.74 & \textbf{163.2} & 70.56 \\
5 & 50 & 4 & 61.00 & 256.00 & 689.20 & 471.20 & 20.58 & 474.80 & 21.52 & \textbf{487.8} & 27.09 \\
5 & 50 & 20 & 33.00 & 461.00 & 323.00 & 160.20 & 58.12 & 164.80 & 58.68 & \textbf{165.4} & 69.01 \\ \hline
10 & 10 & 4 & 45.00 & 362.00 & 685.80 & 447.40 & 18.45 & 399.60 & 16.77 & \textbf{451.6} & 21.54 \\
10 & 10 & 20 & 25.40 & 374.00 & 328.00 & 157.40 & 135.60 & 161.20 & 159.76 & \textbf{161.4} & 80.51 \\
10 & 20 & 4 & 46.60 & 218.02 & 691.60 & 452.40 & 20.29 & 456.60 & 22.86 & \textbf{460.4} & 23.03 \\
10 & 20 & 20 & 26.40 & 368.00 & 328.40 & 154.80 & 67.31 & 160.00 & 68.93 & \textbf{161.4} & 86.91 \\
10 & 50 & 4 & 52.40 & 217.00 & 691.20 & 467.40 & 22.13 & 454.60 & 20.82 & \textbf{481.8} & 30.35 \\
10 & 50 & 20 & 25.80 & 353.00 & 330.60 & 156.60 & 69.87 & 161.60 & 68.25 & \textbf{163.6} & 86.79 \\ \hline 
\bottomrule
\end{tabular}
\end{table}

\end{document}